\documentclass[10pt,journal,compsoc]{IEEEtran}
\ifCLASSOPTIONcompsoc
  \usepackage[nocompress]{cite}
\else
  \usepackage{cite}
\fi
\ifCLASSINFOpdf
\else
\fi

\usepackage{hyperref}       
\usepackage{url}            
\usepackage{booktabs}       
\usepackage{nicefrac}       
\usepackage{microtype}      
\usepackage{subfigure}
\usepackage{makecell}

\usepackage{enumitem}
\usepackage{amsmath}
\usepackage{enumerate}
\usepackage{xcolor}
\usepackage{bm}
\usepackage{graphicx}
\usepackage{color}
\usepackage{chngpage}
\usepackage{import}
\usepackage{amssymb,amsfonts}
\usepackage{textcomp}
\usepackage{flushend}
\usepackage{hhline}
\usepackage{multirow}
\usepackage{array}
\usepackage{gensymb}
\usepackage{nameref}
\usepackage{tabularx}
\usepackage[normalem]{ulem}
\usepackage{algorithm}
\usepackage{algpseudocode}
 \usepackage{hyperref}
\usepackage{ragged2e}
 \DeclareMathOperator{\argmin}{argmin}
\DeclareMathOperator{\argmax}{argmax}
 \usepackage{graphicx}
\usepackage{tikz}
\usepackage{comment} 
\usepackage{amsmath,amssymb} 

\makeatletter
\algrenewcommand\ALG@beginalgorithmic{\small}
\algrenewcommand\algorithmiccomment[2][\small]{{#1\hfill\(\triangleright\) #2}}
\makeatother
\usepackage{enumitem}
\newenvironment{tight_itemize}{
\begin{itemize}[leftmargin=20pt]
  \setlength{\topsep}{0pt}
  \setlength{\itemsep}{0pt}
  \setlength{\parskip}{0pt}
  \setlength{\parsep}{0pt}
}{\end{itemize}}

\DeclareRobustCommand\nobreakspace{\leavevmode\nobreak\ }
\newcolumntype{"}{!{\vrule width 1pt}}
\newcolumntype{L}[1]{>{\raggedright\let\newline\\\arraybackslash\hspace{0pt}}m{#1}}
\newcolumntype{C}[1]{>{\centering\let\newline\\\arraybackslash\hspace{0pt}}m{#1}}
\newcolumntype{R}[1]{>{\raggedleft\let\newline\\\arraybackslash\hspace{0pt}}m{#1}}

\def\eg{\emph{e.g.,}}

\def\etal{\emph{et al.}}
\def\etc{\emph{etc}}
\def\ie{\emph{i.e.,}}

\begin{document}
\title{Gait Recognition in the Wild: A Large-scale Benchmark and NAS-based Baseline}

\author{\normalsize{
Xianda Guo$^*$, Zheng Zhu$^*$, Tian Yang, Beibei Lin, Junjie Huang, Jiankang Deng, \\ Guan Huang, Jie Zhou, Jiwen Lu

\IEEEcompsocitemizethanks{
\IEEEcompsocthanksitem Xianda Guo is with Waytous, Beijing, China. 
\IEEEcompsocthanksitem Tian Yang, Junjie Huang, and Guan Huang are with XForwardAI, Beijing, China. 
\IEEEcompsocthanksitem Zheng Zhu,  Jie Zhou, and Jiwen Lu are with Tsinghua University, Beijing, China. 
\IEEEcompsocthanksitem Jiankang Deng is with Imperial College London, London, UK.
\IEEEcompsocthanksitem Beibei Lin is with the National University of Singapore, Singapore.
\IEEEcompsocthanksitem $^*$ indicates equal contributions. Corresponding author: Jiwen Lu. E-mail: \href{mailto:lujiwen@tsinghua.edu.cn}{lujiwen@tsinghua.edu.cn}.
}
}
}

\markboth{IEEE TRANSACTIONS ON Pattern Analysis and Machine Intelligence, VOL. XX, NO. XX, XXX 2023}%
{Shell \MakeLowercase{\textit{et al.}}: Gait Recognition in the Wild: A Benchmark}

\newcommand{\net}{SPOSGait}

\IEEEtitleabstractindextext{%
\begin{abstract}
\justifying 
Gait benchmarks empower the research community to train and evaluate high-performance gait recognition systems. Even though growing efforts have been devoted to cross-view recognition, academia is restricted by current existing databases captured in the controlled environment. In this paper, we contribute a new benchmark and strong baseline for \textbf{G}ait \textbf{RE}cognition in the \textbf{W}ild (\textbf{GREW}). The GREW dataset is constructed from natural videos, which contain hundreds of cameras and thousands of hours of streams in open systems. With tremendous manual annotations, the GREW consists of 26K identities and 128K sequences with rich attributes for unconstrained gait recognition.  Moreover, we add a distractor set of over 233K sequences, making it more suitable for real-world applications. Compared with prevailing predefined cross-view datasets, the GREW has diverse and practical view variations, as well as more naturally challenging factors. To the best of our knowledge, this is the first large-scale dataset for gait recognition in the wild. Equipped with this benchmark, we dissect the unconstrained gait recognition problem, where representative appearance-based and model-based methods are explored.
The proposed GREW benchmark proves to be essential for both training and evaluating gait recognizers in unconstrained scenarios. In addition, we propose the \textbf{S}ingle \textbf{P}ath \textbf{O}ne-\textbf{S}hot neural architecture
search with uniform sampling for \textbf{Gait} recognition, named \textbf{\net}, which is the first NAS-based gait recognition model. 
In experiments, SPOSGait achieves state-of-the-art performance on the CASIA-B, OU-MVLP, Gait3D, and GREW benchmarks, outperforming existing approaches by a large margin. The code will be released at \url{https://github.com/XiandaGuo/SPOSGait}.

\end{abstract}


}

\maketitle

\IEEEdisplaynontitleabstractindextext

\IEEEpeerreviewmaketitle

\section{Introduction}
\IEEEPARstart{G}ait recognition aims to identify a person according to his/her walking style in a video. Compared with face, fingerprint, iris, and palmprint, gait is hard to disguise and can work at a long distance, giving it unique potential for crime prevention, forensic identification, and social security.
Recognizing gait under a controlled environment has achieved significant progress due to the boom of deep learning. The essential engines of recent gait recognition consist of network architecture evolutions \cite{GLN, GaitPart,takemura2018multi, TSCNN,he2019multi,yu2017invariant, GaitGAN,liao2017pose, GEINet, InputOutput, GaitSet, wu2019spatial, single_gait, nambiar2019gait}, loss function designs \cite{zhang2019cross, TripletLoss, quintuplet, zhang2019gait}, and growing gait benchmarks \cite{USF, UMD, CASIA-B, OU-ISIR, OU-MVLP, OU-ISIR-LP}. Even though gait recognition has achieved impressive advances in past years and possesses the unique advantage of long-distance recognition, this technique has not yet been widely deployed in real-world applications. A notable obstacle is that there is almost no public benchmark to train and evaluate gait recognizers in the wild.

\begin{figure}[htbp]
\centering
\includegraphics[width=1.0\linewidth]{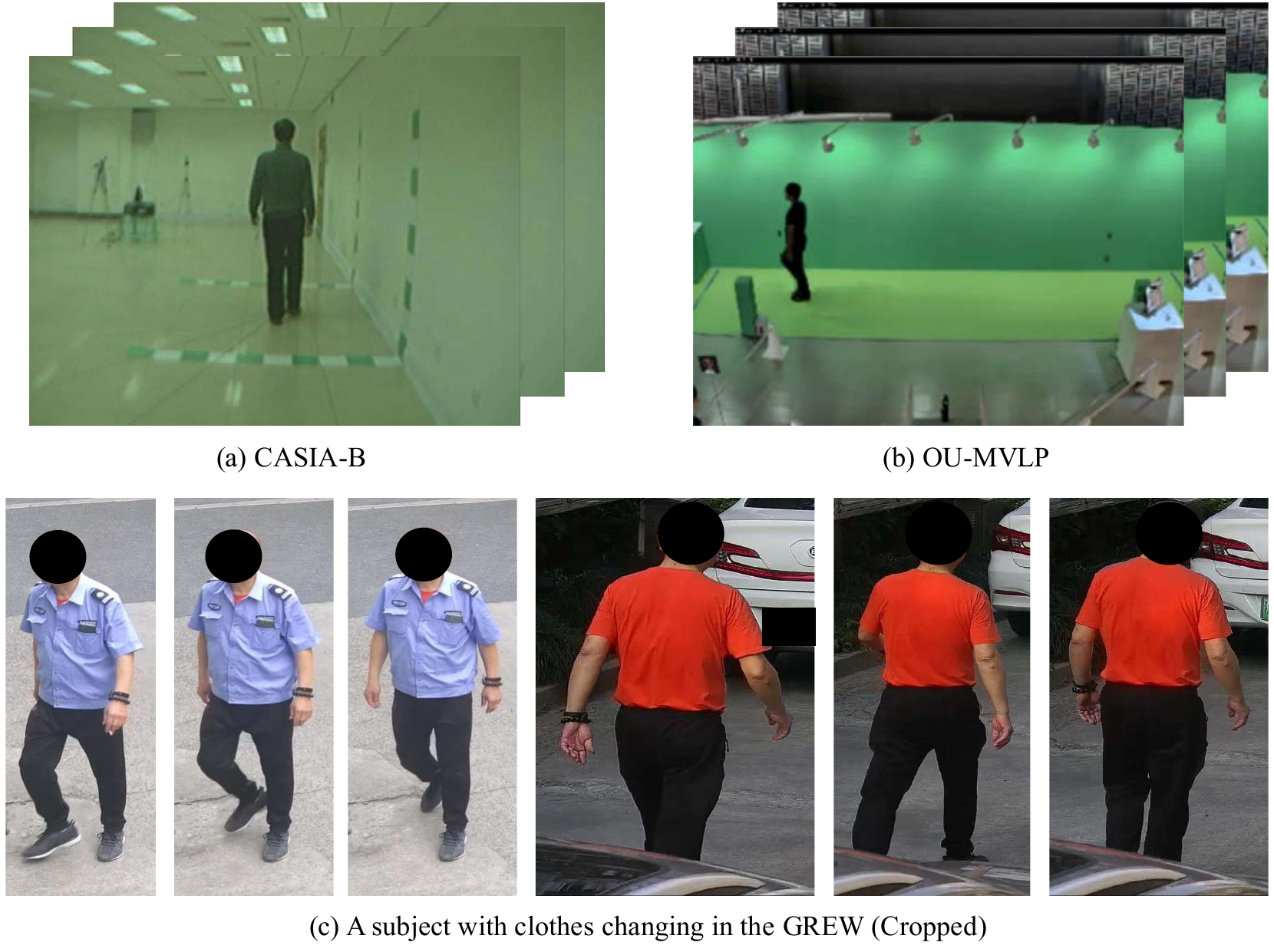}
\vspace{-4mm}
\caption{Examples comparison for CASIA-B \cite{CASIA-B}, OU-MVLP \cite{OU-MVLP} and the proposed GREW. The first two are captured under constrained environments, while the GREW is constructed in the wild. Since OU-MVLP \cite{OU-MVLP} does not release RGB data, visualization results from its original paper are adopted.
Faces are masked in the GREW for privacy concerns.
}
\label{fig:first_sample}
\vspace{-6mm}
\end{figure}

\begin{table*}[htbp]
\caption{\textbf{Top:}Comparison of the GREW with existing gait recognition datasets regarding statistics, data type, captured environment, view variations, and challenging factors. \textbf{Bottom:}Comparison with Video-based and Long-term person re-identification datasets. Datasets are sorted in publication time.
\emph{\#Id.}, \emph{\#Seq.} and \emph{\#Cam.} refer to the numbers of identities, sequences, and cameras. \emph{Sil.}, \emph{Inf.}, \emph{D.} and \emph{A.} mean silhouette, infrared, depth, and audio.
\emph{VI}, \emph{DIS}, \emph{BA}, \emph{CA}, \emph{DR}, \emph{OCC}, \emph{ILL}, \emph{SU}, \emph{SP}, \emph{SH}, and \emph{WD} are abbreviations of view, distractor, background, carrying, dressing, occlusion, illumination, surface, speed, shoes, and walking directions.}
\begin{center}{\scalebox{0.75}{
\begin{tabular}{l|c|c|c|c|c|c|c|c|c}
\hline
Dataset &Publication & \#Id. & \#Seq. & \#Cam. & Data types & \# Distractor & Environment  & View var.   & Challenges \\ \hline
\hline
CMU MoBo~\cite{CMU-Mobo} & TR2001& 25  & 600 & 6 & RGB, Sil. & None & Controlled & Predefined & VI, CA, SP, SU \\
CASIA-A~\cite{CASIA-A} & TPAMI2003& 20 & 240 & 3 & RGB & None & Controlled & Predefined & VI\\
SOTON~\cite{Soton-large} & ASSC2004& 115 & 2,128 & 2 & RGB, Sil. & None & Controlled & Predefined & VI \\
USF ~\cite{USF} & TPAMI2005& 122 & 1,870 & 2 & RGB & None & Controlled & Predefined & VI, CA, SU, SH \\
CASIA-B~\cite{CASIA-B} & ICPR2006& 124 & 13,640 & 11 & RGB, Sil. & None & Controlled & Predefined & VI, CA, DR \\
CASIA-C~\cite{CASIA-C} & ICPR2006& 153 & 1,530 & 1 & Inf., Sil. & None & Controlled & None & CA, SP \\
OU-ISIR Speed~\cite{OU-ISIR-Speed} & CVPR2010& 34 & 612 & 1 & Sil. & None & Controlled & None & SP \\
OU-ISIR Cloth~\cite{OU-ISIR-Clothing} & PR2010& 68 & 2,764 & 1 & Sil. & None & Controlled & None & DR \\
OU-ISIR MV~\cite{OU-ISIR-MV} & ACCV2010& 168 & 4,200  & 25 & Sil. & None & Controlled & Predefined & VI \\
OU-LP~\cite{OU-ISIR-LP} & TIFS2012& 4,007 & 7,842 & 2 & Sil. & None & Controlled & Predefined & VI \\
ADSC-AWD~\cite{ADSC-AWD} & TIFS2014& 20  & 80 & 1 & Sil. & None & Controlled & None & WD \\
TUM GAID ~\cite{GAID} &JVCIR2014 & 305 & 3,370 & 1 & RGB, D., A. & None & Controlled & None & CA, SH \\
OU-LP Age~\cite{OU-ISIR-LP-Age} & CVA2017& 63,846 & 63,846 & 1 & Sil. & None & Controlled & None & Age \\

OU-MVLP~\cite{OU-MVLP} & CVA2018& 10,307 &  288,596  & 14 & Sil. & None & Controlled & Predefined & VI \\
OU-LP Bag~\cite{OU-ISIR-LP-Bag} & CVA2018& 62,528 & 187,584 & 1 & Sil. & None & Controlled & None & CA \\
OU-MVLP Pose~\cite{OU-MVLP-Pose} & TBIOM2020& 10,307 &  288,596  & 14 & 2D Pose & None & Controlled & Predefined & VI \\

VersatileGait~\cite{dou2021versatilegait} & - & 11,000 & 1,032,000 & 33 & Sil. & None & Controlled & Predefined & VI \\
ReSGait~\cite{mu2021resgait} & IJCB2021 & 172 & 870 & 1 & 2D Pose, Sil. & None & Controlled & Predefined & CA, DR, WD \\
BUAA-Duke-Gait~\cite{zhang2022realgait} & - & 1,404 & 4,612 & 8 & RGB, Sil., GEI & None & Wild & Predefined & VI, BA, CA, DR, OCC \\
Gait3D~\cite{zheng2022gait} & CVPR2022& 4,000 &  25,309 & 39 & Sil., 2D/3DPose, 3DMesh & None & Wild & Diverse & VI, SP \\
TTG-200~\cite{liang2022gaitedge} &ECCV2022 &200& 14,198& -&RGB & None &Wild&Diverse &VI,CA,DR,BA \\
SUSTech1K~\cite{shen2023lidargait} &CVPR2023 &1,050& 25,239& 12& RGB, Silhouettes, 3D Point Cloud&None&Wild& Predefined&VI,CA,DR,OCC,ILL,SP,WD\\
GaitLU-1M~\cite{fan2023learning} & TPAMI2023 & 1,035,309 &  1,035,309 & 1,379 & Sil. & None & Wild & Diverse & Real-world \\
\hline
iLIDS-VID~\cite{iLIDS-VID} & ECCV2014 & 300 & 600 & 2 & RGB & None & Wild & Predefined & VI, BA, OCC, ILL \\
MARS~\cite{MARS} & ECCV2016 & 1,261 & 20,715 & 6 & RGB & None & Wild & Predefined & - \\
Duke-Video~\cite{Duke-Video} & CVPR2018 & 1,812 & - & 8 & RGB & None & Wild & Predefined & - \\
Duke-Tracklet~\cite{Duke-Tracklet} & ECCV2018 & 1,788 & - & 8 & RGB & None & Wild & Predefined & - \\
LPW~\cite{LPW} & AAAI2018 & 2,731 & 7,694 & 4 & RGB & None & Wild & Predefined & - \\
LS-VID~\cite{LS-VID} & ICCV2019 & 3,772 & 14,943 & 15 & RGB & None & Wild & Predefined & - \\
CCPG~\cite{li2023depth} &CVPR2023 &  200&16566&10&RGB,Sil&None&Wild&Predefined&VI,CA,DR\\

PRCC  \cite{PRCC} & TPAMI2019 & 221 & - & 3 & RGB & None & Controlled & Predefined & - \\
CVID-reID \cite{CVID-REID} & TMM2020 & 90 & - & - & RGB & None & - & - & - \\
COCAS  \cite{COCAS} & CVPR2020 & 5,266 & - & 30 & RGB & None & Wild & Diverse & OCC, ILL \\

\hline
\textbf{GREW} & - & \textbf{26,345}  & \textbf{128,671} & \textbf{882} & \makecell[c]{\textbf{Sil. Flow}\\\textbf{2/3D Pose}} & \textbf{233,857} & \textbf{Wild} & \textbf{Diverse} & \makecell[c]{\textbf{VI, DIS, BA, CA,} \\ \textbf{DR, OCC, ILL, SU}}\\
\hline
\end{tabular}}}
\end{center}
\vspace{-6mm}
\label{table:training_set}
\end{table*}

To our knowledge, most gait datasets are captured in relatively fixed and constrained environments such as laboratories or static outdoors. CASIA-B \cite{CASIA-B} and OU-MVLP \cite{OU-MVLP} are the most popularly used datasets in recent gait recognition research as shown in Figure \ref{fig:first_sample}. CASIA-B contains 124 subjects and 13,640 sequences, which was constructed in 2006. OU-MVLP consists of 10,307 identities and 288,596 walking videos, making it a big gait dataset with respect to \#subjects. The statistics of more datasets are shown in Table \ref{table:training_set}, which are mainly constructed under controlled settings and designed for predefined cross-view gait recognition. However, in real scenarios, gait recognition would encounter fully unconstrained challenges, such as diverse views, occlusion, various carrying and dressing, complex and dynamic background clutters, illumination, walking style, and surface influence. Existing benchmarks are far behind the requirements of practical gait recognition. Considering the remarkable success of face recognition \cite{Deepface, FaceNet, CosFace, ArcFace, Curricularface, CASIA-WebFace, VGGFace2, MS1M, MegaFace, WebFace260M} and person re-identification (ReID) \cite{AlignedReID, PCB, BoT,xiao2017margin, TripletLoss, HPM,zheng2016person, MARS, Market-1501, DukeMTMC, CUHK03, MSMT17}, it is time to move to benchmark gait recognition in the wild.

\begin{figure}[tbp]
\centering
\includegraphics[width=0.99\linewidth]{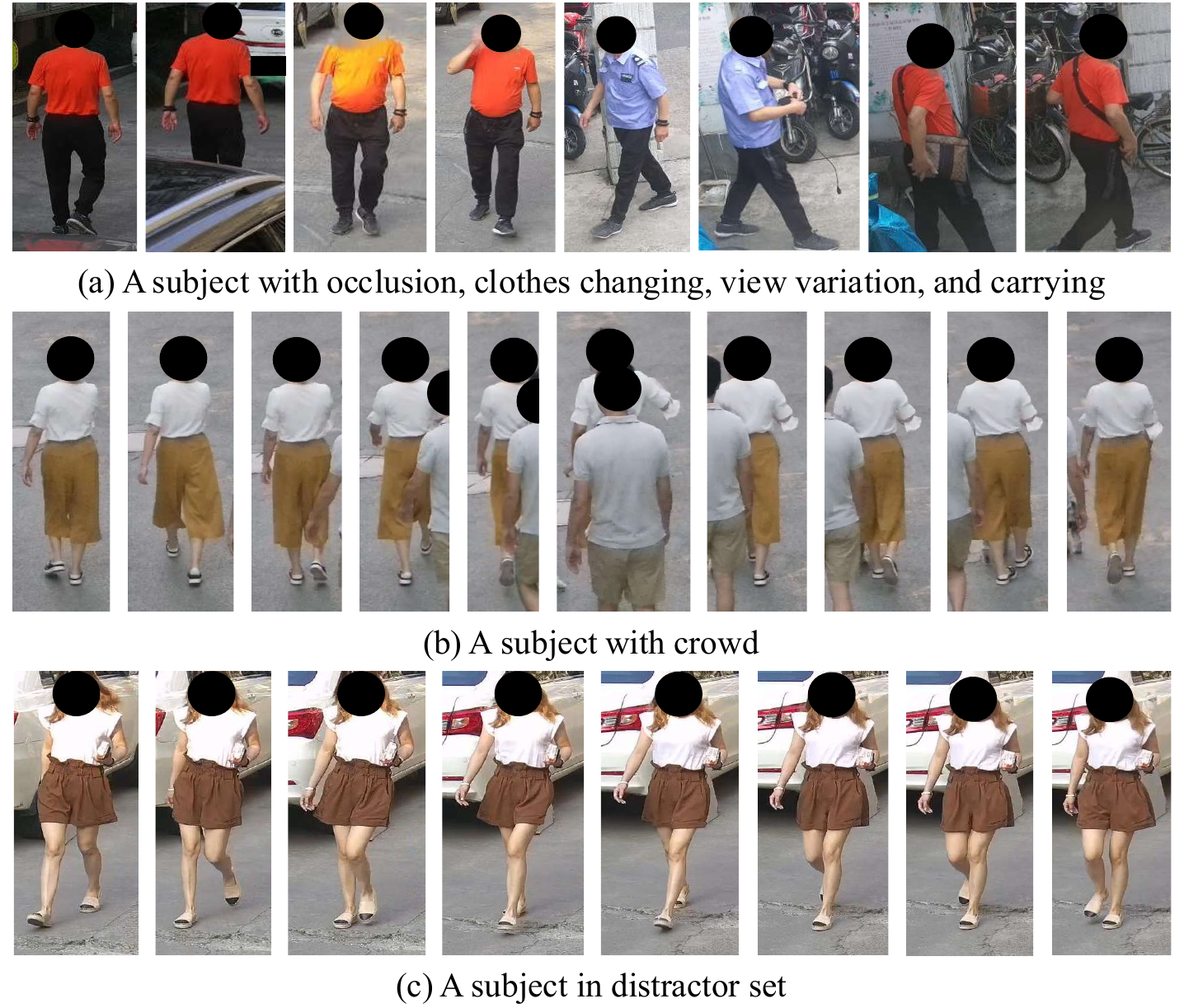}
\vspace{-4mm}
\caption{Identities examples of the GREW dataset. The first two rows show 2 subjects with various challenges. The last row shows a subject from the distractor set. Faces are masked to protect privacy.
}
\label{fig:video_clips}
\vspace{-4mm}
\end{figure}

In this paper, we present the Gait REcognition in the Wild (GREW) benchmark, which is the first work delving into this open problem to the best of our knowledge. The GREW dataset is constructed from natural streams with multiple cameras as shown in Figure \ref{fig:first_sample}. Identity information from raw videos is manually annotated, resulting in 26K subjects, 128K sequences, and 14M boxes for unconstrained gait recognition. Besides, rich human attributes including gender, age group, carrying, and dressing styles are labeled for fine-grained performance analysis. In practice, the gallery scale is a vital problem for recognition accuracy. To this end, we add a distractor set of over 233K sequences, making it more suitable for real-world applications. Since there are a series of gait recognition frameworks using different input data types, the GREW provides silhouettes, Gait Energy Images (GEIs)~\cite{GEI}, optical flow, and 2D/3D poses by automatical processing. Compared with controlled gait datasets such as CASIA-B and OU-MVLP, our GREW is fully unconstrained and contains more diverse and practical view variations instead of predefined ones. Meanwhile, there are various challenging factors in the GREW such as large distractor set, complex background, occlusion, carrying, and dressing as shown in Table \ref{fig:first_sample} and Figure \ref{fig:video_clips}.



To equip the proposed benchmark with a strong baseline, we leverage the power of NAS to automate the process of network architecture design and search for the optimal network components for gait recognition. To the best of our knowledge, this is the first attempt to apply NAS to the task of gait recognition in the wild. 
Specifically, we propose the \textbf{S}ingle \textbf{P}ath \textbf{O}ne-\textbf{S}hot neural architecture search with uniform sampling for \textbf{Gait} recognition, called \textbf{\net}. By doing so, we establish a strong baseline model that can significantly improve the state-of-the-art performance for gait recognition in the wild. 

With the proposed GREW benchmark and SPOSGait baseline, the unconstrained gait recognition problem is deeply investigated. Firstly, representative appearance-based and model-based approaches are performed on the GREW, which indicates a lot of room for improvement. With the distractor set, gait recognition in the wild would become more challenging. 
Meanwhile, SPOSGait
achieves state-of-the-art performance on the GREW benchmark, outperforming existing approaches by a large margin.
Then, the influence of the data scale is explored, including the number of training identities and gallery size. Increasing training subjects consistently boosts the performance, while the large-scale test set with distractors is still difficult for CNN-based recognizer. Lastly, performances on different attributes (gender, age group, carrying, and dressing) are reported, which gives in-depth analysis results. 

The main contributions can be summarized as follows:
\begin{tight_itemize}
\item A large-scale benchmark is constructed for the research community toward gait recognition in the wild. The proposed GREW consists of 26K subjects and 128K sequences with rich attributes from flexible data streams, which makes it the first dataset for unconstrained gait recognition to the best of our knowledge.

\item To constitute the GREW benchmark, we collect thousands of hours of streams from multiple cameras in open systems. With automatical pre-processing and tremendous manual identity annotations, there are more than 14M boxes that simultaneously provide silhouettes and human poses. Besides, we enrich the GREW with a distractor set with 233K sequences, making it more suitable for real-world applications.

\item Enabled by the new benchmark, we perform extensive gait recognition experiments, including representative methods, scale influences, attributes analysis, and pre-training. Results indicate that the GREW is necessary and effective for gait recognition in the wild. Besides, recognizing unconstrained gait is a very challenging task for current approaches.

\item We present the first NAS-based framework for gait recognition, namely SPOSGait, which achieves state-of-the-art performance on the CASIA-B, OU-MVLP, Gait3D, and GREW benchmark. The network structure identified by SPOSGait on the GREW dataset, when retrained on OU-MVLP and Gait3D, consistently achieves state-of-the-art results, highlighting its remarkable effectiveness and adaptability.
We hope this work can inspire further research in this area and contribute to the development of more effective and efficient gait recognition systems.

\end{tight_itemize}

In this extended version for the TPAMI submission, we have made significant improvements to our work, which was initially published as a preliminary conference version in ICCV 2021~\cite{grew}. The enhancements are as follows: (1) We propose a novel gait recognition method based on NAS. This approach leverages the power of NAS to automatically discover optimal network architectures for gait recognition tasks, potentially enhancing the overall performance and robustness of the model. We aim to push the boundaries of gait recognition research and contribute to the state of the art in the field. (2) SPOSGait achieves state-of-the-art performance on the CASIA-B, OU-MVLP, Gait3D, and GREW benchmark. (3) We have expanded our experiments by including the latest methods GaitGL~\cite{gaitgl} and CSTL~\cite{cstl}, leading to a more exhaustive comparison. By evaluating the performance of our method against these approaches, we demonstrate the superiority of the proposed baseline in various aspects. (4) We provide an in-depth literature review encompassing the field of gait recognition, as well as a comprehensive analysis of pertinent benchmarks.

Since the preliminary version of this work was published in ICCV 2021~\cite{grew}, the GREW benchmark has been significantly influencing the gait recognition community, fostering further research, inspiring new methodologies, and promoting the development of practical applications~\cite{wang2022gaitstrip,wang2023dygait,zhang2022realgait,segundo2023long,zheng2022gait,fan2022opengait,mu2021resgait}.  

\section{Related Works}

\subsection{Gait Recognition Benchmarks}

CMU MoBO~\cite{CMU-Mobo}, CASIA-A~\cite{CASIA-A}, OU-ISIR Speed~\cite{OU-ISIR-Speed}, OU-ISIR Cloth~\cite{OU-ISIR-Clothing} and ADSC-AWD~\cite{ADSC-AWD} contain less than 100 identities. The CASIA-A~\cite{CASIA-A} and ADSC-AWD~\cite{ADSC-AWD} are relatively small datasets with only 20 identities. The CMU MoBO~\cite{CMU-Mobo} dataset was created in 2001 which includes 25 identities and provides silhouette masks and bounding boxes. The OU-ISIR Speed~\cite{OU-ISIR-Speed} includes 34 individuals captured in lateral vision, with speeds ranging from 2 to 10 km/h in 1 km/h intervals, which are useful for studying actions with limited variations in body shapes and speeds, respectively. Meanwhile, the OU-ISIR Cloth~\cite{OU-ISIR-Clothing} is useful for studying the impact of clothing on action recognition, which is composed of 68 people in lateral vision with 32 clothing variations.

SOTON~\cite{Soton-large}, USF~\cite{USF}, CASIA-B~\cite{CASIA-B}, CASIA-C~\cite{CASIA-C}, OU-ISIR MV~\cite{OU-ISIR-MV}, TUM GAID~\cite{GAID}, and ReSGait~\cite{mu2021resgait} include more identities than above-mentioned datasets. The SOTON~\cite{Soton-large} dataset is the first one to investigate the effect of elapsed time on gait recognition. Comprising 122 subjects, the USF dataset~\cite{USF} captures participants walking under a variety of conditions, including diverse surfaces, angles, and velocities. Meanwhile, the TUM GAID~\cite{GAID} database combines audio, image, and depth~\cite{hofmann2014tum} to study multimodal gait recognition. In comparison, the CASIA-C dataset~\cite{CASIA-C} captures 153 subjects via infrared cameras (thermal spectrum) in four distinct scenarios: normal walking, slow walking, fast walking, and normal walking with a backpack. Despite having fewer subjects, the CASIA-B~\cite{CASIA-B} dataset is widely used for evaluating gait recognition algorithms because of its early release and high-quality gait images that exhibit variations in body poses, walking speeds, and clothing.

OU-LP~\cite{OU-ISIR-LP}, OU-LP Age~\cite{OU-ISIR-LP-Age}, OU-MVLP~\cite{OU-MVLP}, OU-MVLP Bag~\cite{OU-ISIR-LP-Bag}, OU-MVLP Pose~\cite{OU-MVLP-Pose}, VersatileGait~\cite{dou2021versatilegait}, BUAA-Duke-Gait~\cite{zhang2022realgait}, Gait3D~\cite{zheng2022gait}, and GaitLU-1M~\cite{fan2023learning} consist of more than 1000 identities. The OU-LP series~\cite{OU-ISIR-LP-Age, OU-ISIR-LP-Bag, OU-MVLP, OU-MVLP-Pose} have been expanded to include comprehensive variants such as different carrying bags~\cite{OU-ISIR-LP-Bag}, subjects of different ages~\cite{OU-ISIR-LP-Age}, and annotations of 2D pose~\cite{OU-MVLP-Pose}. Because of its large population, the OU-MVLP~\cite{OU-MVLP} dataset has gained popularity in the current research community. To automatically generate a large-scale synthetic gait dataset, the VersatileGait~\cite{dou2021versatilegait} uses a game engine, which includes approximately one million silhouette sequences of 11,000 subjects with fine-grained attributes in various complex scenarios. Recently, LidarGait~\cite{shen2023lidargait}
explores 3D gait features from point clouds and proposes LidarGait.

\textbf{Comparison with other gait datasets in the wild}. 
Following the publication of our ICCV 2023 paper~\cite{grew}, Gait3D~\cite{zheng2022gait} provides dense 3D information on body shapes, viewpoints, and dynamics from recovered 3D SMPL models in video frames. 
Meanwhile, GaitLU-1M~\cite{fan2023learning} dataset collects a large-scale unlabeled gait dataset, which consists of 1.02 million walking sequences. In comparison, the GREW dataset distinguishes itself with its extensive manual annotations and a strong focus on practical, real-world scenarios. Its comprehensive coverage of diverse environments and real-life variations provides a more robust and applicable dataset for real-world gait recognition challenges, highlighting its unique advantage in the field.

\textbf{Comparison with Video-based and Long-term Person ReID}. Another related computer vision task is person ReID in the videos and long-term (cloth changing) ReID. Gait recognition approaches aim to identify a certain subject by silhouettes (GEIs) or pose information, instead of RGB input in ReID. This feature makes the gait recognizer more friendly for preserving privacy, which may be more easily accepted by the public. Recently, CCPG~\cite{li2023depth} presents a comprehensive analysis of person re-identification and gait recognition, specifically focusing on the challenging scenario of varying clothing conditions, offering novel insights and solutions for robust identification in dynamic real-world environments. Additionally, it is noteworthy that gait patterns are inherently more difficult to conceal or disguise. 
Moreover, compared with popular video ReID ~\cite{iLIDS-VID, MARS, Duke-Video, Duke-Tracklet, LPW, LS-VID} and long-term ReID ~\cite{CVID-REID, COCAS, PRCC} datasets, our GREW has more \#identities and \#cameras as shown in Table~\ref{table:training_set}.

\subsection{Gait Recognition Approaches}

\textbf{Model-based} methods~\cite{PoseGait, GaitGraph, li2021end, zheng2022gait, li2022multi} take the estimated underlying structure of the human body as input, such as 2D/3D pose and SMPL model. Among them, GaitGraph~\cite{GaitGraph} is a recent model-based approach for gait recognition with promising results on the CASIA-B dataset. It utilizes a combination of 2D human pose input and graph convolutional network to achieve gait recognition. 
Li \etal~\cite{li2021end}introduces a novel cross-view gait recognition method that synchronizes multi-view gait sequences during training to extract consistent pose sequences for testing. 
PoseGait~\cite{{PoseGait}} leverages 3D human pose as input for gait recognition, which is estimated using ~\cite{3D_pose}. ~\cite{li2020end} firstly employs an SMPL model in gait recognition. SMPLGait~\cite{zheng2022gait} explores 3D human meshes based on the SMPL model for gait recognition, which is a novel 3D gait recognition framework. However, model-based methods are typically inferior to appearance-based methods in performance comparison due to inaccurate pose estimation from low-quality images and the absence of identity-related shape information.

\textbf{Appearance-based} methods~\cite{chai2022lagrange, xu2023end, GaitSet, GaitPart, cstl, wang2023dygait, fan2022opengait} typically encode pedestrian images and subsequently identify individuals based on the acquired gait embeddings. The Gait Energy Image (GEI) method~\cite{GEI} projects a sequence of gait silhouettes into a single image, resulting in a significant reduction in computational cost. However, the GEI-based methods~\cite{GEI, GEINet, TSCNN} often sacrifice discriminative power for this efficiency gain. With the advent of deep learning, most current appearance-based approaches primarily concentrate on spatial feature extraction and the modeling of gait temporal dynamics. Among them, GaitSet~\cite{GaitSet} is one of the most influential works in recent years. Specifically, GaitSet~\cite{GaitSet} represents gait sequence assets of frames, which are subsequently processed using deep learning techniques to extract discriminative features. GaitPart~\cite{GaitPart} focuses on the analysis of body parts for effective gait recognition and employs a micromotion capture module to model the temporal dependencies effectively. GaitGL~\cite{gaitgl} proposes a Global and Local Feature Extractor (GLFE) to effectively captures both global and local feature using 3D convolution. CSTL~\cite{cstl} introduces a Context-Sensitive Temporal Feature Learning (CSTL) network, which enables the extraction of motion representation by considering the temporal contextual information. 
Chai ~\etal~\cite{chai2022lagrange} introduces a novel motion extraction module and a view-embedding module, enhancing the performance in cross-view gait recognition.
OpenGait~\cite{fan2022opengait} revisits the recent developments in gait recognition and proposes a simple but powerful baseline model, called GaitBase. In general, it remains a challenging task to design an effective structure for gait recognition, even with the significant efforts made by the research community.

\subsection{Neural Architecture Search}

In recent years, NAS has achieved significant success in reducing the human effort required for designing neural networks in various high-level vision tasks, such as classification~\cite{cai2018proxylessnas, wu2019fbnet, fang2019densely}, object detection~\cite{tan2019efficientdet, peng2019efficient} and semantic segmentation~\cite{liu2019auto, zhang2019customizable, chen2020fasterseg}. Furthermore, recent research has extended NAS techniques to more specialized domains, such as architecture search for person re-identification~\cite{quan2019auto,li2021combined}, face recognition~\cite{zhu2019neural}, and stereo matching~\cite{saikia2019autodispnet, cheng2020hierarchical}.



Although NAS has achieved significant success in these fields, it has not yet been applied to gait recognition. Therefore, there is a need for further research to investigate the effectiveness of NAS in this field and to develop novel approaches that can overcome the challenges specific to gait recognition in the wild. To address this gap, we draw inspiration from Guo~\etal~\cite{guo2020single}, who proposed an innovative Single Path One-Shot (SPOS) model designed to overcome obstacles arising during the training phase by creating a simplified supernet with single-path architectures. We propose to apply the SPOS model to gait recognition research and establish a strong baseline model in this paper. By leveraging NAS techniques in gait recognition, we can potentially reduce the manual effort required in designing effective gait recognition models while achieving state-of-the-art performance.


\section{The GREW Dataset}


\subsection{Overview of GREW}

\begin{figure}[htbp]
\centering
\includegraphics[width=0.95\linewidth]{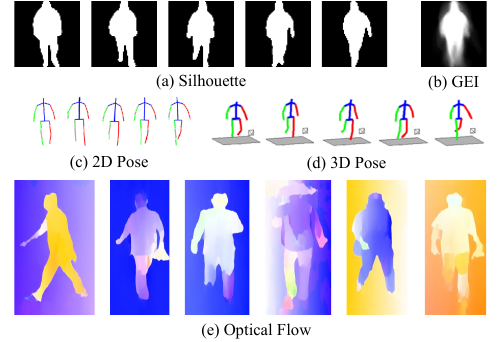}
\vspace{-5mm}
\caption{Examples of silhouette, GEI, 2D and 3D human pose and optical flow from the GREW dataset.}
\label{fig:input_training_set}
\vspace{-6mm}
\end{figure}

Qualitative and quantitative comparisons between the GREW and representative gait recognition datasets are illustrated in Figure \ref{fig:first_sample} and Table \ref{table:training_set}, respectively. The GREW consists of 26,345 subjects and 128,671 sequences, which come from 882 cameras in open environments. Furthermore, we propose the first distractor set in the gait research community, which contains 233,857 sequences. Silhouettes, GEIs, and 2D/3D human pose data types are provided for both appearance-based and model-based algorithms as shown in Figure \ref{fig:input_training_set}.
Since the raw data is captured in natural environments, recognizing identities by gait in the GREW is more challenging compared with popular CASIA-B and OU-MVLP. For example, detecting and segmenting the human body from the complex and dynamic background is a difficult task, considering occlusion, truncation, illumination, \etc. As shown in Figure \ref{fig:video_clips}, unconstrained settings also bring new challenging factors for gait patterns, such as diverse views, dressing, carrying, crowd, and distractors.

\begin{figure}[htbp]
\centering
\includegraphics[width=1.0\linewidth]{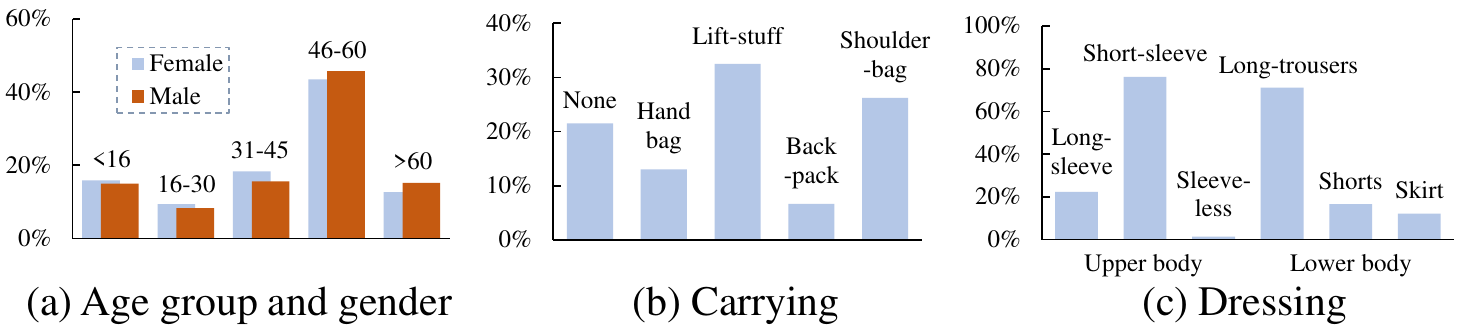}
\vspace{-6mm}
\caption{Age group, gender, carrying and dressing attributes in the GREW. In (c), upper body dressing styles contain long-sleeve, short-sleeve, and sleeveless, while lower body includes long-trousers, shorts, and skirts.}
\label{fig:human_attributes}
\vspace{-6mm}
\end{figure}

\subsection{Data Collection and Annotation}

The raw videos are collected from 882 cameras in large public areas, during one day in July 2020. About 70\% cameras have
non-overlapping views, and all cameras cover more than
600 positions. \emph{We are authorized by administrations, and all of the involved subjects are told to collect data for research purposes}.
7,533 video clips are used, containing nearly 3,500 hours of 1080$\times$1920 streams.

Before annotation, HTC detector \cite{HTC} is performed to provide initial human boxes. Then, annotators select the boxes from the same subject as a trajectory (sequence). Since there are multiple cameras and a certain person may enter/leave the same camera view, one identity always has multiple sequences. We ensure that each subject in the GREW \texttt{train}, \texttt{val}, and \texttt{test} set appears at more than 1 camera, which guarantees view diversity.
Other sequences are utilized as distractor sets as shown in Section \ref{sec_distractor}.

In Table \ref{table:training_set}, we compare GREW with previous gait datasets regarding \#identities, \#sequences, \#cameras, provided data types, \#distractor set, environment, view variations, and challenging factors.
Finally, a total of 128,671 sequences are manually annotated to obtain 26,345 identities, which contain 14,185,478 human boxes.
Besides, the distractor set consists of 233,857 sequences and 9,676,016 human boxes.
It takes 20 annotators to work for 3 months for this tremendous labeling, and we hope the proposed GREW benchmark will facilitate future research on unconstrained gait recognition. \emph{It is worth noting that only silhouettes, optical flow, and poses (shown in \autoref{fig:input_training_set}) will be utilized and released}. In the data collection phase, conspicuous notifications, such as signboards, were prominently displayed at the site to inform all individuals that the data gathered would be used exclusively for research purposes. Additionally, we have implemented a protocol whereby any individual captured in the video data who wishes to have their personal data removed can request deletion by contacting us via email.


\subsection{Automatical Pre-processing}

Representative gait recognition approaches can be roughly divided into appearance-based \cite{GEINet, TSCNN, GaitSet, GaitPart,li2020gait, li2020end, GLN} and model-based \cite{GaitGraph, PoseGait, OU-MVLP-Pose,li2019joint,liu2016memory} categories, which take silhouettes (GEIs) and human poses as input, respectively.
In the GREW benchmark, we provide both two data types by automatical pre-processing. Specifically, silhouettes are produced by segmenting the foreground human body utilizing the HTC \cite{HTC} algorithm. We also try the Mask R-CNN \cite{maskrcnn}, which results in inferior gait recognition accuracy.
It is worth noting that human detection and segmentation may be less accurate as shown in Figure \ref{fig:input_training_set}. Compared with near-perfect results of CASIA-B and OU-MVLP in the static background, the GREW enables assessing the influence of less heuristic pre-processing for gait recognition. This is a topic of great interest for practical applications but rarely considered in previous datasets. For GEIs, we do not adopt the gait cycle due to imperfect detection and segmentation in the wild.
For human pose estimation, we provide 2D and 3D key points by \cite{HRNet} and \cite{3D_pose} as illustrated in Figure \ref{fig:input_training_set}. Furthermore, optical flow \cite{flownet2, flownet2_code} is extracted for potential usage as shown in Figure \ref{fig:input_training_set}.


\begin{figure}[htbp]
\centering
\includegraphics[width=0.99\linewidth]{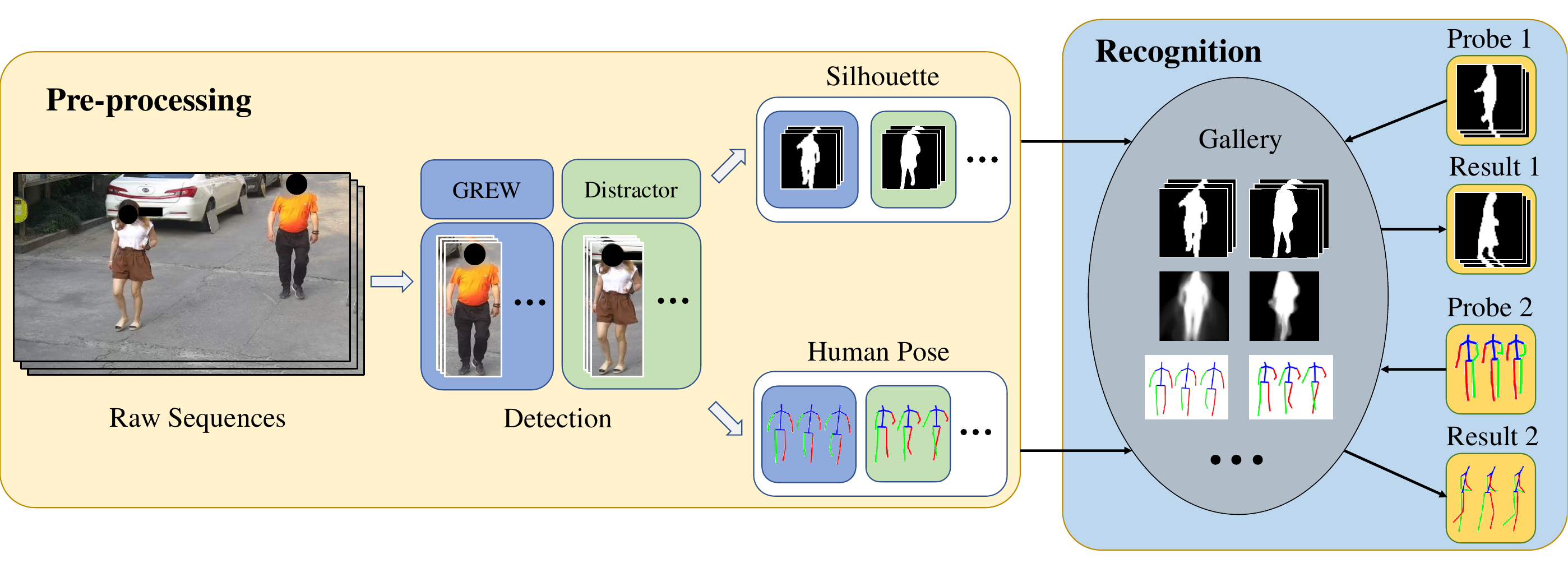}
\vspace{-4mm}
\caption{The pipeline of gait recognition in the wild, consisting of pre-processing and recognition steps. The pre-processing part detects humans from raw sequences and provides silhouette or pose information. Given a certain probe, the recognition part performs 1:N searching from the gallery.}
\label{fig:pipeline}
\vspace{-6mm}
\end{figure}

\subsection{Human Attributes}

For fine-grained recognition analysis, we annotate each sequence with rich attributes. Soft biometric features including gender and age are labeled for all subjects. Ages are categorized into 5 groups, which adopt 14-year intervals for adults (\ie 16 to 30, 31 to 45, 46 to 60). Children (under 16) and elders (over 60) are treated as separate groups. The statistics of gender and age group are given in Figure \ref{fig:human_attributes}. For each age group, there is an almost balanced male and female distribution. Since carrying and dressing are influential for gait pattern extraction, the GREW benchmark further provides 5 carrying conditions (\ie none, backpack, shoulder bag, handbag, and lift-stuff) and 6 dressing styles (\ie upper-long-sleeve, upper-short-sleeve, upper-sleeveless, lower-long-trousers, lower-shorts, and lower-skirt). Detailed statistics of these attributes are illustrated in Figure \ref{fig:human_attributes}. Subjects in more than 70\% sequences carry something, while upper-short-sleeves and lower-long-trousers form the majority of cloth styles.

\subsection{Distractor Set}
\label{sec_distractor}

In real-world applications of gait recognition, the gallery scale is a vital factor. Therefore, we further augment the GREW benchmark with an additional distractor set.
This dataset contains 233,857 sequences and 9,676,016 boxes, consisting of extra walking trajectories not belonging to the GREW \texttt{train}, \texttt{val}, and \texttt{test}. Specifically, identities that are labeled but only appear at 1 camera would be categorized into the distractor set. In Section \ref{Exp}, apart from the GREW \texttt{test} set, we also report results on the GREW \texttt{test} + \texttt{distractor} set.

\subsection{Evaluation Protocol}

The GREW dataset is divided into 3 parts: a \texttt{train} set with 20,000 identities and 102,887 sequences, a \texttt{val} set with 345 identities and 1,784 sequences, a \texttt{test} set with 6,000 identities and 24,000 sequences. Identities in 3 sets are captured in different cameras.
Each subject in \texttt{test} set has 4 sequences, 2 for probe and 2 for gallery. Besides, there is a distractor set with 233,857 sequences. Detailed statistics of the splits are presented in Table \ref{table:statistics_splits}.

\begin{table}[htbp]
\vspace{-4mm}
\caption{Statistics of different splits.}
\vspace{-6mm}
\centering
\begin{center}{
\smallskip
\begin{tabular}{c|c|c|c}
\hline
Split & \#Identities  & \#Sequences & \#Frames \\
\hline\hline
Train & 20,000  & 102,887 & 10,166,842  \\
Val & 345  & 1,784 & 238,532  \\
Test & 6,000  & 24,000 & 3,780,104  \\
Distractor & 233,857  & 233,857 & 9,676,016  \\
\hline
\end{tabular}}
\end{center}
\label{table:statistics_splits}
\vspace{-4mm}
\end{table}

As shown in Figure \ref{fig:pipeline}, in the inference stage, recognizing gait in the wild firstly detects the subject from raw videos. Then, the segmentation or pose estimation module is performed to obtain gait input. Gait recognition is always a 1:N searching process, which aims to retrieve the same person from the gallery given a probe subject. When evaluated on the \texttt{test} set, the gait probe and gallery are all paired. When evaluated on a certain attribute, a subset of probes (sequences with the corresponding attribute) are chosen to perform gait recognition. We adopt prevailing Rank-$k$ as the evaluation metric, which denotes the possibility of locating at least one true positive in the top-$k$ ranks.

\section{Gait Recognition Network Architecture Search}

\begin{figure*}[htbp]
\centering
\includegraphics[width=0.8\textwidth]{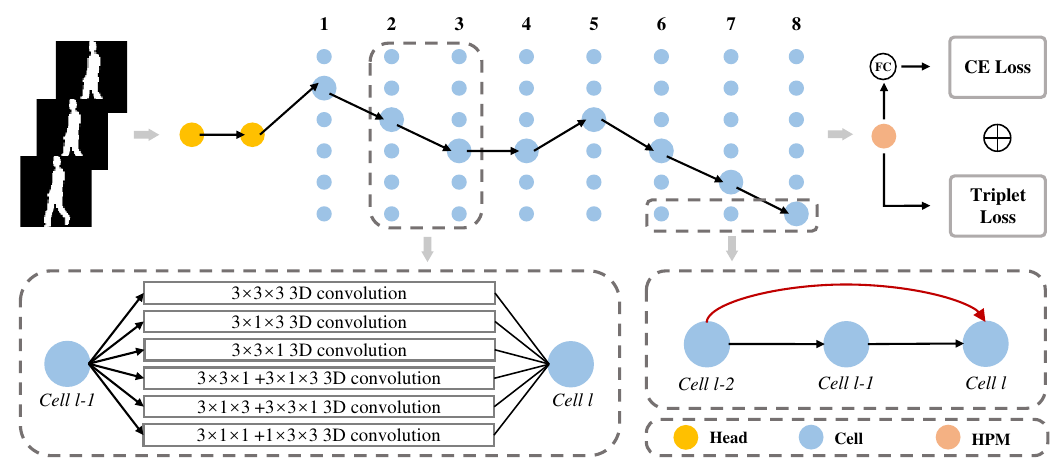}
\vspace{-4mm}
\caption{\textbf{Overview of the proposed SPOSGait.}  FC denotes the Fully Connected Layer, while HPM represents Horizontal Pyramid Mapping\cite{GaitSet}. The head is composed of $3 \times 3 \times3$ 3D convolutional layers, which are followed by a Batch Normalization layer and a ReLU activation layer.}
\vspace*{-2em}
\label{fig:overview}
\end{figure*}

In this section, we present a novel SPOS-based NAS approach for gait recognition, as shown in Figure~\ref{fig:overview}. Our proposed approach leverages prior human knowledge and successful handcrafted designs in the field, enabling efficient and effective search for the optimal architecture. Our approach achieves state-of-the-art performance in gait recognition, surpassing previous results.

The search space for network architectures, denoted as $\mathcal{A}$, is represented by a \emph{supernet}, denoted as $\mathcal{N}(\mathcal{A}, W)$, where $W$ refers to the weights of the supernet. In order to carry out an architecture search, we utilize the one-shot approach with uniform sampling~\cite{guo2020single}. Firstly, the optimization of the supernet weight is conducted as
\begin{equation}
    W_\mathcal{A} = \mathop{\argmin}_W \mathcal{L}_\text{train} \left( \mathcal{N}(\mathcal{A}, W) \right).
\label{equ:supernet_oneshot}
\end{equation}
Secondly, the architecture search is executed as
\begin{equation}
\begin{split}
    a^* = \mathop{\argmax}_{a\in \mathcal{A}} &\text{ACC}_\text{val} \left( \mathcal{N} (a, W_\mathcal{A}(a)) \right).\\
\end{split}
\label{equ:arch_search_oneshot}
\end{equation}
During the search process, each sampled architecture $a$ inherits its weights from $W_\mathcal{A}$, denoted as $W_\mathcal{A}(a)$. Given that the architecture weights are already accessible, the evaluation of $\text{ACC}_{val}(\cdot)$ solely relies on inference, rendering the search remarkably \emph{efficient}. The search is further characterized by its \emph{flexibility}, as it accommodates any appropriate search algorithm. Moreover, the search can be performed multiple times on the trained supernet. Lastly, upon completion of the architecture search process, the final step is retraining the discovered optimal architecture for enhanced performance.

\subsection{Supernet Training}
\label{subsection:Supernettraing}

\begin{figure}[htbp]
\centering
\includegraphics[width=0.7\linewidth]{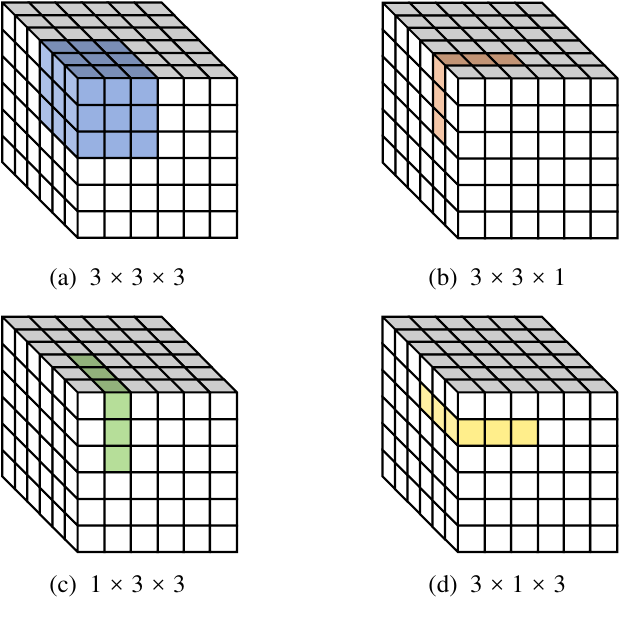}
\vspace{-4mm}
\caption{Visualization of Feature Extraction in candidate 3D convolutional networks.}
\label{fig:candidate}
\vspace{-6mm}
\end{figure}

In this section, we focus on employing a single-path supernet in conjunction with uniform sampling for an efficient and flexible neural architecture search process. The single-path supernet is designed to encompass multiple subnetworks, sharing the same structure. However, at each layer, multiple alternative operations, such as different types of convolutions, are available. This design enables the construction of each subnetwork by selecting a specific operation at each layer, effectively reducing computational and storage overhead as all subnetworks share the same weights and parameters.

Uniform sampling serves as the sampling strategy in our architecture search process, dictating how architectures are chosen from the search space for training and evaluation. This approach involves uniformly selecting architectures from the search space, assigning equal probability to each architecture. The advantage of this method lies in its simplicity and independence from prior knowledge or complex search strategies. 
By combining the single-path supernet with uniform sampling, we can efficiently explore and evaluate different neural network architectures in the search space. This approach helps to reduce computational costs while maintaining the flexibility and generality of the search process.

\textbf{Candidate Operation Selection.} To ensure that our proposed NAS-based approach for gait recognition can effectively capture the discriminative features of gait sequences, we carefully select a set of candidate operators that can perform 3D convolutions in the neural network. As shown in \autoref{fig:candidate}, the set of candidates operators includes $\mathcal{O}^F =\{$ ''$3 \times 3 \times3$ 3D convolution'', ''$3 \times 1 \times3$ 3D convolution'', ''$3 \times 3 \times1$ 3D convolution'',''$3 \times 3 \times1$ +$3 \times 1 \times3$ 3D convolution'',''$3 \times 1 \times3$ +$3 \times 3 \times1$ 3D convolution'',''$3 \times 1 \times1$ +$1 \times 3 \times3$ 3D convolution''$\}$. Each 3D convolution layer integrates Batch Normalization~\cite{ioffe2015batch} and ReLU~\cite{nair2010rectified} activation mechanisms. These candidate operators represent different ways to perform 3D convolutions in the neural network. They vary in terms of kernel sizes and combinations, which can influence the network's capacity to learn features from the input data. Since gait recognition is a complex problem that involves both spatial and temporal features, we select candidate operators that have characteristics suitable for capturing both types of features. For instance, some operators may be better at capturing spatial information, while others may excel in learning temporal patterns that are crucial for gait recognition. This careful selection of candidate operators is crucial for the success of our proposed NAS-based approach for gait recognition.


\textbf{Residual Connection.} 
Inspired by the residual connection architecture of ResNet \cite{he2016deep}, we propose an approach to the commonly used direct cell design for constructing cell outputs, as noted in previous research \cite{liu2018darts,liu2019auto,saikia2019autodispnet,cheng2020hierarchical}. As shown in the bottom right corner of Figure~\ref{fig:overview}, the residual connection, highlighted by a red line, is implemented using $1\times1\times1$ 3D convolution followed by a Batch Normalization~\cite{ioffe2015batch} layer.
This inclusion allows the network to learn residual mappings alongside direct mappings.

\textbf{Data Augmentation}
To align with practical applications, we investigate a Data Augmentation (DA) strategy tailored for gait silhouettes. While data augmentation has been widely used in computer vision applications, few studies have investigated its effectiveness for gait recognition. Recently, OpenGait~\cite{fan2022opengait} introduced five distinct data augmentation techniques (including Horizontal Flip, Rotation, Perspective Transformation, Affine Transformation, and Random Erasing) specifically tailored for gait recognition. While OpenGait~\cite{fan2022opengait} conducted ablation experiments solely on the Gait3D~\cite{zheng2022gait} and CASIA-B~\cite{CASIA-B} datasets, we extend this research by providing comprehensive ablation experiments on the GREW dataset. Our work aims to unify and optimize the data augmentation strategy within the GREW dataset context.



\begin{table}[htbp]
\centering
\vspace{-4mm}
\caption{Searched Network Structure. ''c'' represents the channels of each layer's output and ``c\_last'' refers to the channel of the last layer. ``GeM'' refers to Generalized-Mean pooling~\cite{gaitgl} and ``bin'' represents the scales of GeM. ``BNNeck'' is from~\cite{fan2022opengait}.}
\label{tab:searchnetwork}
\scalebox{0.88}{
\begin{tabular}{l|c|c}
\hline
\multicolumn{3}{c}{\textbf{CASIA-B}}\\ 
\hline
\textbf{Name} &  \textbf{Layer setting} & \textbf{Input} \\ \hline
Layer0 & Conv3d ($ 3 \times 3 \times 3 $), 96    &   Silhouettes  \\ 
\hline
Layer1& Conv3d($ 1 \times 1 \times 1 $)  &  Silhouettes          \\
\hline
Downsample0  &\begin{tabular}[c]{@{}c@{}}MaxPool3d \\ k=(1, 2, 2) s=(1, 2, 2)\end{tabular}                  & Layer0+Layer1  \\ 
\hline

\hline
Layer2   &  Conv3d($ 1 \times 1 \times 1 $ )  &  Downsample0 \\ 
\hline

Layer3 &  Conv3d($3 \times 1 \times 3 $+$ 3 \times 3 \times 1 $), 192  & Layer2  \\ 
\hline
Layer4   &  Conv3d($ 1 \times 1 \times 1 $ )&  Layer2 \\ \hline
Downsample1   & \begin{tabular}[c]{@{}c@{}}MaxPool3d \\ k=(1, 2, 2) s=(1, 2, 2)\end{tabular}&    Layer3 + Layer4  \\ 
\hline

Layer5 & Conv3d($ 3 \times 3 \times 3 $), 256  &Downsample1 \\ 
\hline
Layer6  &Conv3d($ 1 \times 1 \times 1 $)    &      Downsample1              \\ 
\hline

Layer7 & Conv3d($ 3 \times 1 \times 3 $), 256     & Layer5+Layer6\\ 
 \hline
Layer8   & Conv3d($ 1 \times 1 \times 1 $ )    &   Layer5+Layer6   \\ 
\hline

Layer9   & Conv3d($ 1 \times 1 \times 1 $ ) &  Layer7+Layer8\\ 
\hline
GeM   & GeM,bin=[16]                   &               Layer9        \\ 
\hline
Head & FC, 256, part=16 &GeM \\
\hline
BNNecks&\begin{tabular}[c]{@{}c@{}}FC, class\_num=74\\part=16\end{tabular} &Head\\
\hline \hline

\multicolumn{3}{c}{\textbf{GREW, Gait3D, OU-MVLP}}\\ 
\hline
\textbf{Name} &  \textbf{Layer setting} & \textbf{Input} \\ \hline
Layer0\_0 & Conv3d ($ 3 \times 3 \times 3 $), c[0]    &   Silhouettes  \\ 
\hline
Layer0\_1 & Conv3d ($ 3 \times 3 \times 3 $), c[0]   &    Layer0\_0   \\
\hline
Layer1& Conv3d($ 1 \times 1 \times 1 $)  &  Silhouettes          \\
\hline
Downsample0  &\begin{tabular}[c]{@{}c@{}}MaxPool3d \\ k=(1, 2, 2) s=(1, 2, 2)\end{tabular}                  & Layer0\_1+Layer1  \\ 
\hline

Layer2\_0 &  Conv3d($3 \times 1 \times 3 $+$ 3 \times 3 \times 1 $), c[1]  & Downsample0  \\ 
\hline
Layer2\_1 & Conv3d($ 3 \times 1 \times 3 $+$ 1 \times 3 \times 3 $), c[1] & Layer2\_0 \\ 
\hline
Layer3   &  Conv3d($ 1 \times 1 \times 1 $ )&  Downsample0 \\ \hline

Layer4   & Conv3d( $ 3\times 1 \times 1 $), c[1]   &     Layer2\_1 + Layer3   \\ 
\hline

Layer5\_0 & Conv3d($ 3 \times 3 \times 1 $), c[2]   & Layer4 \\ 
\hline
Layer5\_1 & Conv3d($ 3 \times 3 \times 3 $), c[2]   &     Layer5\_0            \\ 
\hline
Layer6   &  Conv3d($ 1 \times 1 \times 1 $ )  &  Layer4 \\ 
\hline
Downsample1   & \begin{tabular}[c]{@{}c@{}}MaxPool3d \\ k=(1, 2, 2) s=(1, 2, 2)\end{tabular}&     Layer5\_1+   Layer6  \\ 
\hline

Layer7\_0 & Conv3d($ 3 \times 1 \times 3 $+$ 3 \times 3 \times 1 $), c[3]  &Downsample1 \\ 
\hline
Layer7\_1 & Conv3d($ 3 \times 3 \times 1 $+$ 3 \times 1 \times 3 $), c[3] &Layer7\_0  \\ 
\hline
Layer8  &Conv3d($ 1 \times 1 \times 1 $)    &         Downsample1              \\ 
 \hline

Layer9\_0 & Conv3d($ 3 \times 1 \times 3 $+$ 3 \times 3 \times 1 $), c[3]     & Layer7\_1 + Layer8\\ 
 \hline
Layer9\_1 & Conv3d($ 3 \times 3 \times 1 $+$ 3 \times 1 \times 3 $), c[3]  &Layer9\_0 \\ 
 \hline
Layer10   & Conv3d($ 1 \times 1 \times 1 $ )    &   Layer8   \\ 
\hline

Layer11   & Conv3d($ 1 \times 1 \times 1 $ ) &  Layer9\_1+Layer10\\ 
\hline
GeM   & GeM,bin                  &               Layer11        \\ 
\hline

Head & FC, c\_last, part &GeM \\
\hline
BNNecks&\begin{tabular}[c]{@{}c@{}}FC, class\_num, part\end{tabular} &Head\\
\hline

\end{tabular}
}
\vspace{-6mm}
\end{table}

\subsection{Evolutionary Architecture Search and Retraining} 
\label{subsection:Search&Retraining}



We employ the evolutionary algorithm~\cite{guo2020single} for our search process, in which each architecture is only utilized for inference. This approach is highly efficient. The NAS algorithm's goal is to search for the optimal combination of these candidate operators to construct a neural network that performs well on the given task. During the search process, the algorithm will explore different configurations of these operators and evaluate their performance. Based on the evaluation results, the algorithm will find the best combination of candidate operators.

After completing the architecture search process, the final step is retraining the discovered optimal architecture. This step is essential for obtaining the best performance from the selected architecture, as it refines the weights and biases specific to the chosen network configuration. During the retraining phase, the entire network is trained from scratch, using the optimal architecture found during the search process. This stage typically involves using the same training dataset, along with an updated set of hyperparameters such as learning rate, batch size, and regularization factors. The retraining process ensures that the model can achieve the best possible performance. As a result, the retrained model is expected to yield improved performance compared to its initial state during the architecture search.

\subsection{Loss Function and Optimization}
To achieve the best performance, we employ two different loss functions to optimize our model: triplet loss and cross-entropy loss~\cite{GaitSet}.
During the supernet Training phase in Section~\ref{subsection:Supernettraing}, we only employ triplet loss to enhance the efficiency of training. During the Retraining phase in Section~\ref{subsection:Search&Retraining}, our model is optimized by combining triplet loss and cross-entropy loss to learn discriminative feature representations while effectively handling classification tasks. The combined loss function can be represented as
\begin{equation}
\label{combined_loss}
Loss_{all} = \lambda_1 L_\text{triplet} + \lambda_2 L_\text{CE},
\end{equation}
where $L_\text{triplet}$ and $L_\text{CE}$ mean triplet loss and cross-entropy loss respectively. $\lambda_1$ and $\lambda_2$ are scalar weight parameters that balance the contributions of triplet loss ($L_{\text{triplet}}$) and cross-entropy loss ($L_{\text{CE}}$) to the combined loss function $Loss_{all}$.







\subsection{Searched Network Structure}

We initially conducted a network architecture search on the GREW dataset. The network structure identified through this search on GREW could be directly applied to retrain the OU-MVLP and Gait3D datasets, achieving state-of-the-art results. For the CASIA-B dataset, considering its relatively smaller size, a correspondingly smaller network layer structure is required. Therefore, it necessitated a separate network architecture search specifically for CASIA-B. The structure identified through this dedicated search is presented in Table~\ref{tab:searchnetwork}.
In the GREW dataset, the network configuration comprises a channel sequence of 64, 128, 256, and 512. It accommodates a total of 20,000 classes. The bin settings are arranged as 16, 8, 4, 2, and 1, with the partition count set to 31 and the terminal channel count fixed at 512. In the case of the Gait3D dataset, the channel configuration remains the same. However, it is tailored for 3,000 classes with a uniform bin setting of 16. The partitioning is designed for 16 parts, and the terminal channel count is adjusted to 256. For the OU-MVLP dataset, the channel sequence is identical to that of Gait3D, supporting 5,153 classes, with the bin setting and partition count mirroring that of Gait3D, and the terminal channel count also set to 256.

\textbf{Analysis} As detailed in Table~\ref{tab:searchnetwork}, the diversity in the configuration of these Conv3d layers, with varied combinations like $3 \times 1 \times 3$, $3 \times 3 \times 1$, and straightforward $1 \times 1 \times 1$, is particularly noteworthy. This variety indicates a strategic approach to dissect and analyze the spatial-temporal dimensions of gait data. The interleaving of downsampling layers after complex convolutional combinations points to an efficient extraction and compression of gait features. This structure allows the network to distill essential information while reducing computational load, crucial for processing high-dimensional gait data. Overall, this NAS-optimized architecture offers significant insights into the effective use of 3D convolutions for gait recognition. It underscores the potential of varying kernel sizes and layer configurations in understanding the complex dynamics of human gait, providing a valuable blueprint for future exploration in this field. The structure reflects a keen understanding of the spatial-temporal nature of gait data and suggests pathways for further research to enhance the robustness and accuracy of gait recognition systems, especially in diverse and uncontrolled environments.

\section{Experiments}

\subsection{Main Evaluation Results}
\label{main_results}
In this section, we reproduce six existing appearance-based methods and two model-based methods. We also perform a comprehensive analysis between our proposed baseline and these methods.

\subsubsection{Appearance-based Approaches}
\noindent{\bf GEINet}~\cite{GEINet} directly learns gait representation features from GEIs and then corresponds
to identities. The network of the GEINet has 4 layers, consisting of 2 convolutions and 2 Fully-Connected (FC) layers. Softmax loss is adopted for optimization, and output from the last FC is utilized to calculate the distance between the probe and the gallery.

\noindent{\bf TS-CNN}~\cite{TSCNN} framework adopts two-stream CNN architecture which learns similarities between GEIs pair for gait recognition. MT architecture setting is utilized in this paper, which matches mid-level features at the top layer. TS-CNN also takes GEIs as input and has 6 layers. 2-class Cross-entropy loss is used for training, while the classifier
indicates the probability of two subjects whether they are the same one during inference.

\noindent{\bf GaitSet}~\cite{GaitSet} uses several convolution and pooling layers to extract convolutional templates on an unordered silhouette set. Batch All triplet loss \cite{TripletLoss}
is adopted for optimizing, and 15,872-$d$ embedding features are utilized for recognition during inference. Following the OU-MVLP training setting, we use more channels convolutional layers and 250K iterations with 2 learning rate schedules.

\noindent{\bf GaitPart}~\cite{GaitPart} proposes a part-based network design focusing on fine-grained representation and micro-motion in different parts of the human body. Training and testing on the GREW benchmark follow most GaitSet settings.

\noindent{\bf CSTL}~\cite{cstl} employs multi-scale learning on the temporal dimension of the sequence. This approach enables the model to learn both long-term and short-term motion dynamics, which are crucial for accurate gait recognition. Training and evaluation protocols are maintained consistent with those in the original paper.

\noindent{\bf GaitGL}~\cite{gaitgl} employs 3D CNN to learn
both global and local features from gait sequences. Training and testing on the GREW benchmark follow most GaitGL settings.

\subsubsection{Model-based Approaches}


\noindent{\bf PoseGait}~\cite{PoseGait} explores 3D human pose as gait recognition input which is estimated by \cite{3D_pose}. And 2D pose extracted from \cite{HRNet} is utilized to obtain 3D pose information. For the gait feature part, a 22-layer (20 convolutions and 2 FC) CNN with 512-$d$ embedding is trained for extraction, which is optimized by Softmax and Center losses.


\noindent{\bf GaitGraph}~\cite{GaitGraph} is a recent model-based gait recognition approach with a promising result on CASIA-B.
This work combines 2D human pose input and graph convolutional network to achieve gait recognition. Supervised contrastive loss is utilized to optimize the graph network, and we strictly follow its augmentation and training details. During the evaluation, the 256-$d$ feature vector is extracted for calculating the distance between the probe and the gallery.

\subsubsection{Implementation Details}
Excluding SPOSGait, all appearance-based and model-based methods are re-implemented in one codebase using PyTorch \cite{PyTorch} and trained on a cluster (each with 8 $\times$ 2080TI GPUs, Intel E5-2630-v4@2.20GHz CPU, 256G RAM). For all methods, the input resolution is 64 $\times$ 44. For re-implemented approaches, we train models for 250K iterations with a batch size of ($p=32$, $k=4$). The learning rate starts at $10^{-4}$ and decreases to $10^{-5}$ after 150K iterations. SPOSGait is implemented on a cluster (each with 8 $\times$ 3090 GPUs, 96 $\times$ Intel(R) Xeon(R) Platinum 8255C CPU @ 2.50GHz, 376G RAM). For the proposed SPOSGait, the SGD optimizer is employed with an initial learning rate of 0.1 and a weight decay of 0.0005. The input frame number is fixed as 30. The batch size is ($p=32$, $k=4$). We set the triplet loss margin as 0.2. 

\noindent{\bf Supernet training stage} For GREW, the total number of iterations is 500,000, and the learning rate is reduced at milestones 250,000 and 350,000. Only triplet loss is employed with a margin value of 0.2. Moreover, we use the entire GREW \texttt{train} set for training. 
Eight NVIDIA RTX 3090 GPUs are utilized to train our SPOSGait model, requiring approximately three days.
For CASIA-B, the total number of iterations is 100,000, and the learning rate is reduced at milestones 40,000, 60,000, and 80,000. Both triplet loss and cross-entropy loss are used. This model, also trained on eight NVIDIA RTX 3090 GPUs, is completed within a remarkably efficient timeframe of 14 hours.

\noindent{\bf Search stage} We set the population size to $P=50$, the maximum number of iterations to $\mathcal{T}=20$, and $Top_{k}$ to 10.  

\noindent{\bf Retraining stage} For GREW, the total number of iterations is 200,000, and the learning rate is reduced at milestones [50,000, 100,000, 150,000]. The triplet loss and cross-entropy loss are used. In addition, we utilize spatial data augmentation techniques, including random horizontal flipping and rotation.
For Gait3d, the total number of iterations is 70,000, and the learning rate is reduced at milestones [20,000, 40,000, 50,000, 60,000]. The triplet loss and cross-entropy loss are used. Additionally, data augmentation techniques such as Perspective Transformation, Horizontal Flip, and Rotation are applied. The batch size is set to ($p=32$, $k=4$) to ensure effective training. 
For OU-MVLP, the total number of iterations is 120000, and the learning rate is reduced at milestones [60,000, 80,000, 100,000]. The triplet loss and cross-entropy loss are used. The batch size for OU-MVLP is configured as ($p=32$, $k=8$), considering the dataset's specific characteristics. 

\subsubsection{Performance Comparisons}

\begin{figure}[htbp]
\centering
\vspace{-4mm}
\includegraphics[width=0.8\linewidth]{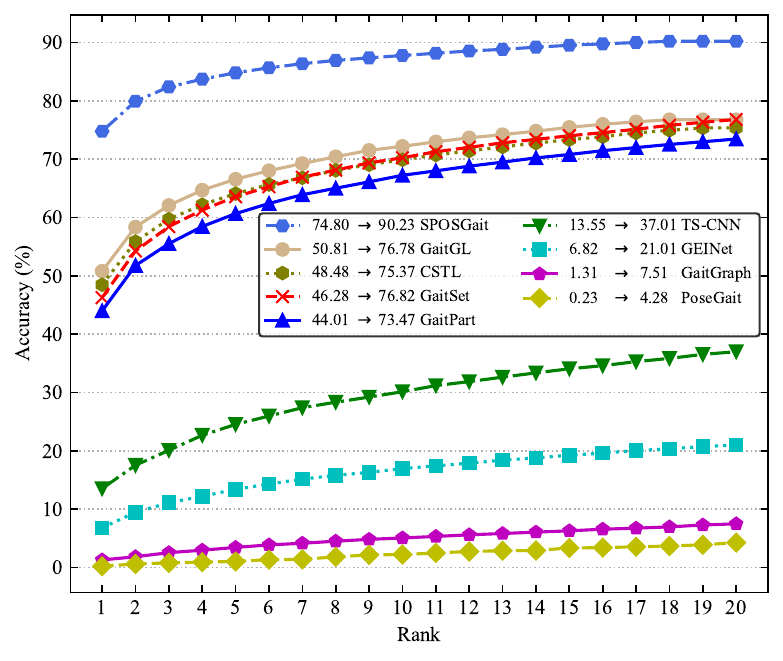}
\vspace{-4mm}
\caption{Rank-$k$ results (\%) of baselines. Trained on the GREW \texttt{train} and evaluated on \texttt{test} set. Legends show Rank-1$\rightarrow$ Rank-20 accuracies.}
\label{fig:rank_k}
\vspace{-4mm}
\end{figure}

\begin{table}[htbp]
\caption{Rank-1, Rank-5, Rank-10, Rank-20 (\%) on the GREW dataset. Trained on the GREW \texttt{train} set and evaluated on \texttt{test} set. The models marked with $^\dagger$ indicate the results of the original paper. The models marked without $^\dagger$ indicate our implementation.}
\vspace{-4mm}
\centering
\begin{center}{\scalebox{0.99}{
\smallskip
\begin{tabular}{l|c|c|c|c}
\hline
Method & Rank-1 &  Rank-5  & Rank-10 & Rank-20 \\
\hline\hline
PoseGait~\cite{PoseGait} & 0.23  & 1.05 & 2.23 & 4.28\\
GaitGraph~\cite{GaitGraph} & 1.31 & 3.46 & 5.08 & 7.51 \\
GEINet~\cite{GEINet} & 6.82 & 13.42 & 16.97 & 21.01 \\
TS-CNN~\cite{TSCNN} & 13.55 & 24.55 & 30.15 & 37.01 \\
GaitSet~\cite{GaitSet} & 46.28 & 63.58 & 70.26 & 76.82 \\
GaitPart~\cite{GaitPart} & 44.01 & 60.68 & 67.25 & 73.47 \\
CSTL~\cite{cstl} & 48.48 & 64.12  & 69.91 & 75.37\\
GaitGL~\cite{gaitgl} & 50.81 &66.58  &  72.22 &76.78 \\
\hline
MTSGait \cite{zheng2022gaitmulti}$^\dagger$ & 55.32  & 71.28  & 76.85 &  81.55 \\
RealGait \cite{zhang2022realgait}$^\dagger$ & 54.12  & 71.47 & 77.57  & 81.71  \\
GaitBase \cite{fan2022opengait}$^\dagger$ & 60.1  & -  & -  & -  \\
GaitGCI \cite{dou2023gaitgci}$^\dagger$ & 68.5& 80.8& 84.9 &87.7  \\
DyGait \cite{wang2023dygait}$^\dagger$ & 71.4 &83.2 &86.8 &89.5  \\
\hline
SPOSGait &\textbf{74.80} &\textbf{84.82} &\textbf{87.79}  &\textbf{90.23} \\
\hline
\end{tabular}}}
\end{center}
\label{table:rank_k}
\vspace{-6mm}
\end{table}

We compare the proposed strong baseline SPOSGait with
other published state-of-the-art methods, as illustrated summarized in Table \ref{table:rank_k}. The Rank-$k$ accuracies are also shown in Figure \ref{fig:rank_k}. The GREW \texttt{train} and \texttt{test} sets are utilized for training and evaluation, respectively. 
Compared to these methods, SPOSGait consistently achieves the highest Rank-1 accuracy. This is particularly notable in diverse datasets such as CASIA-B, OU-MVLP, Gait3D, and GREW (both test and test + distractor sets). The architecture identified through SPOSGait on the GREW dataset exhibits remarkable adaptability. Upon retraining, this architecture consistently achieves state-of-the-art results on the OU-MVLP and Gait3D datasets. This demonstrates the robustness of SPOSGait and its exceptional versatility. Such cross-dataset generalization is seldom seen in gait recognition research. The ability of SPOSGait to maintain superior performance, even when transitioning between datasets with only minimal modifications and retraining, is indicative of its advanced design. This quality underscores SPOSGait's potential as a universally applicable solution in the evolving landscape of gait recognition, capable of adapting to a variety of challenges and settings, thereby affirming its leading status across multiple advanced datasets.

\begin{table}[htbp]
\vspace{-4mm}
\caption{Rank-1 accuracy (\%) comparison on CASIA-B, OU-MVLP, GREW \texttt{test}, GREW \texttt{test} + \texttt{distractor}. Reported results on CASIA-B and OU-MVLP are averaged accuracies excluding identical-view cases (averaged again on NM, BG, and CL for CASIA-B). OU-MVLP does not provide RGB data, so there is no result for model-based methods.
}
\vspace{-5mm}
\centering
\begin{center}{\scalebox{0.8}{
\smallskip
\begin{tabular}{l|c|c|c|c|c}
\hline
Baseline & \makecell[c]{CASIA-B\\($64 \times 44$)}  &  OU-MVLP &Gait3D & GREW \texttt{test} & \makecell[c]{GREW \texttt{test} \\ + \texttt{distractor}} \\
\hline\hline
PoseGait~\cite{PoseGait} & 41.38 & - &-&0.23  &0.18 \\
GaitGraph~\cite{GaitGraph} & 76.26 & - &-& 1.31 & 0.66 \\
GEINet~\cite{GEINet} & 68.71 & 31.47 & 5.40&6.82 & 4.21 \\
TS-CNN~\cite{TSCNN} & 72.06 & 34.72 & -&13.55 & 10.41\\
GaitSet~\cite{GaitSet} & 83.64 & 86.97&36.7 & 46.28 & 41.97 \\
GaitPart~\cite{GaitPart} & 87.51 & 88.36&28.2 & 44.01 & 39.95 \\
CSTL~\cite{cstl} & 91.86& 90.2& 11.70&48.48&44.32 \\
GaitGL~\cite{gaitgl} & 91.83 & 89.7 &29.7&50.81&46.67\\
GaitBase~\cite{fan2022opengait}& 89.66 & 90.8&64.6 &60.1&-\\
\hline
SPOSGait &\textbf{92.12}&\textbf{92.26} &\textbf{64.94}&\textbf{74.80}  &\textbf{68.56} \\

\hline
\end{tabular}}}
\end{center}
\label{table:comparison_casiab_mvlp_grew}
\vspace{-4mm}
\end{table}

Considering that the GREW is the first unconstrained gait benchmark, we compare the results on CASIA-B and OU-MVLP, and GREW datasets, as shown in Table~\ref{table:comparison_casiab_mvlp_grew}. It is important to note that the compared models exhibit near 90\% Rank-1 accuracy on the CASIA-B and OU-MVLP datasets, yet their performance declines to approximately 50\% on the GREW dataset. This highlights the challenges of real-world gait recognition and suggests that the GREW dataset is essential and effective for evaluating unconstrained gait recognition, with ample room for improvement. The SPOSGait model demonstrates significant advantages in GREW \texttt{test} and GREW \texttt{test} + \texttt{distractor} settings. 


\subsection{Ablation Study of SPOSGait}

In this part, we perform ablation studies during the retraining phase to evaluate the effectiveness of Loss, BatchNorm, Residual Connection, and Data Augmentation in the proposed SPOSGait.

\begin{table}[htbp]
\vspace{-4mm}
\caption{Ablation of loss. Trained on the GREW \texttt{train} set and evaluated on \texttt{test} set.}
\vspace{-4mm}
\centering
\begin{center}{
\smallskip
\begin{tabular}{l|c|c|c}
\hline
Method&Loss & Rank-1 & Epoch \\
\hline\hline
\multirow{5}{*}{\centering SPOSGait} &triplet loss & 62.54 & 20\\
&cross-entropy loss& 74.41 & 20\\
&1.0*triplet+1.0*cross-entropy & 72.75& 20 \\
&1.0*triplet+0.1*cross-entropy &70.14 & 20 \\
&0.1*triplet+1.0*cross-entropy &\textbf{74.80} & 20 \\
\hline
\end{tabular}}
\end{center}
\label{table:ablation_loss}
\vspace{-4mm}
\end{table}

\noindent\textbf{Impact of Loss.}
Table~\ref{table:ablation_loss} demonstrates the performance of various combinations of triplet loss and cross-entropy loss on the GREW dataset with the Rank-1 accuracy, which is the primary evaluation metric. When only using the triplet loss, the model achieves a Rank-1 accuracy of 62.54\%. By contrast, only employing the cross-entropy loss yields a higher Rank-1 accuracy of 74.41\%. Interestingly, when the weight of the triplet loss is decreased (0.1*triplet + 1.0*cross-entropy), the model achieves the highest Rank-1 accuracy of 74.80\%. This finding suggests that the cross-entropy loss plays a more significant role in the model's performance, and appropriately balancing the contribution of the triplet loss is crucial for achieving optimal results. 
This effect is partially attributed to the GREW dataset being collected in uncontrolled scenarios, where subjects' gait trajectories are not regulated, offering fewer perspectives compared to controlled gait data. Additionally, another factor involves the handling of different sequences for the same ID; when sequences are excessively long, we opt to split the sequence, leading to a relatively small intra-class distance. This specific handling elevates the effectiveness of cross-entropy loss.

\begin{table}[htbp]
  \centering
  \caption{Trained on the GREW \texttt{train} set and evaluated on \texttt{test} set. BN indicates BatchNorm. Skip indicates Skip connection. DA indicates Data Augmentation. HF refers to Horizontal Flip, R refers to Rotation, PT refers to Perspective Transformation, AT refers to Affine Transformation, and RE refers to Random Erasing. Settings used in our final model are underlined.}
  \begin{tabular}{l|c c|ccccc|c}
    \hline
    \multirow{2}{*}{Method} & \multirow{2}{*}{\textbf{BN}} & \multirow{2}{*}{\textbf{Skip}} & \multicolumn{5}{c|}{\textbf{Data Augmentation}}& \multirow{2}{*}{Rank-1} \\
    & & &HF&R&PT&AT&RE&\\
    \hline \hline
    \multirow{9}{*}{\centering SPOSGait} & \checkmark & & & & & &&53.37 \\
    & \checkmark & \checkmark & & & & & & 70.05 \\
    & \checkmark & \checkmark & \checkmark & & & & &  73.69\\
    & \checkmark & \checkmark &  &\checkmark & & & &  73.02\\
    & \checkmark & \checkmark &  & &\checkmark & & &  73.27\\
    & \checkmark & \checkmark &  & & & \checkmark& &66.59  \\
    & \checkmark & \checkmark &  & & & & \checkmark& 70.58 \\
    & \checkmark & \checkmark & \checkmark &\checkmark & & & & \underline{74.80} \\
    & \checkmark & \checkmark & \checkmark &\checkmark & \checkmark& & & \textbf{74.97} \\
    \hline
  \end{tabular}
  \label{table:Ablation}
\end{table}


\noindent\textbf{Impact of BatchNorm.}
The BatchNorm layer can effectively facilitate model convergence, as shown in Table~\ref{table:Ablation}, thus promoting training stability. In SPOSGait, it is difficult to train the network without the incorporation of BatchNorm layers. The BatchNorm layer in the SPOSGait plays a crucial role in ensuring a stable and efficient training process.

\noindent\textbf{Impact of Residual Connection.}
As shown in Table~\ref{table:Ablation}, incorporating skip connections into the model results in a substantial increase in Rank-1 accuracy, reaching 70.05\%. Residual connections facilitate the efficient flow of gradients throughout the network during backpropagation, mitigating the vanishing gradient issue and empowering the model to learn more effectively.

\noindent\textbf{Impact of Data Augmentation.}
As shown in Table~\ref{table:Ablation}, the inclusion of Horizontal Flip increases Rank-1 accuracy to 73.69\%, demonstrating its effectiveness in enhancing generalization. Similarly, the addition of Rotation results in a comparable accuracy of 73.02\%, indicating its value for model robustness. Perspective Transformation slightly outperforms Rotation with a 73.27\% accuracy, highlighting the importance of capturing perspective variations. In contrast, Affine Transformation reduces accuracy to 66.59\%, suggesting potential overfitting or alignment issues, while Random Erasing yields a moderate improvement to 70.58\%. The integration of Horizontal Flip, Rotation, and Perspective Transformation notably achieves the highest Rank-1 accuracy of 74.97\%, pointing to the complementary benefits of these techniques. This finding suggests that data augmentation effectively mitigates over-fitting issues arising from noisy variations typically encountered in real-world applications.

\subsection{More Analysis of the GREW Dataset}

In this section, we perform more experiments and analyses on the proposed GREW dataset. Firstly, we investigate the influence of the scale including increasing training and testing identities, and distractor set size. Secondly, performance on different human attributes is compared, consisting of accuracy on gender, age group, carrying condition, and dressing style. Last comes sample results on successes and failures of gait recognition.

\subsubsection{Influence of the Scale}
\label{Exp}

In the deep learning era, large-scale labeled data plays a significant role in bench-marking various vision tasks \cite{ImageNet, COCO, MS1M, Market-1501}.
In this section, we investigate the data scale influence for training and testing on the GREW.

\begin{figure}[htbp]
\centering
\includegraphics[height=0.8\linewidth]
{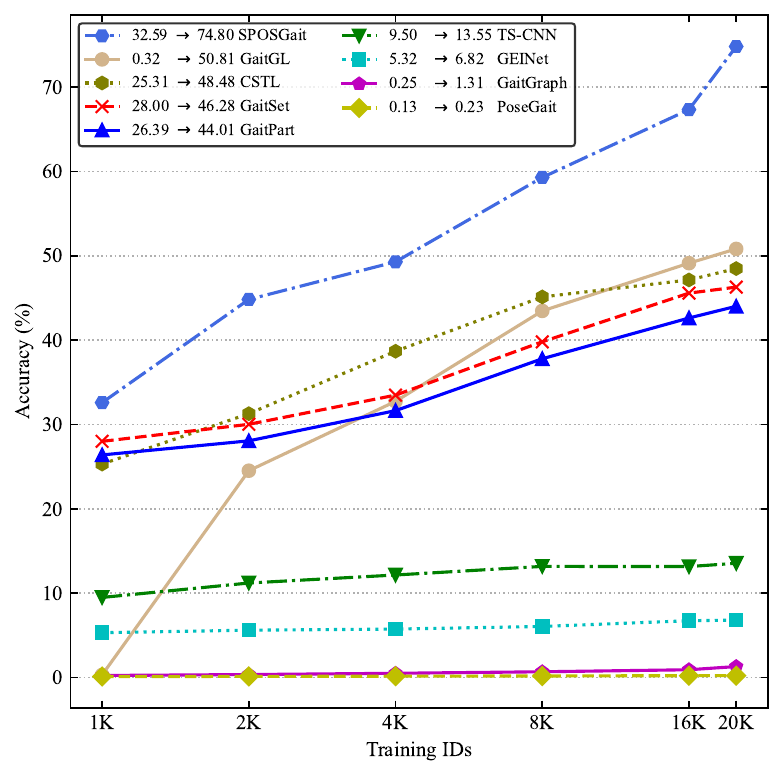}
\vspace{-5mm}
\caption{Rank-1 accuracy (\%) on \texttt{test} set with increasing training identities.
The legend shows performance changes from 1K to 20K data.
}
\label{fig:increasing_training_identities}
\vspace{-3mm}
\end{figure}

\noindent{\bf Accuracy with Increasing Training Identities.}
In this experiment, we demonstrate gait recognition accuracy with increasing training identities. 6 different subset sizes are prepared, including 1K, 2K, 4K, 8K, 16K, and a maximum of 20K. The first 5 training subsets are randomly chosen but fixed for different algorithms. The evaluation is performed on the whole GREW \texttt{test} set. As presented in Figure \ref{fig:increasing_training_identities}, for GaitGL and SPOSGait, the Rank-1 on \texttt{test} set grows stably with more training identities. Therefore, the 20K size of the whole training set achieves the highest Rank-1 accuracy. Specifically, SPOSGait increases the Rank-1 from 32.59\% on 1K training subjects to 74.80\% on 20K subjects. The results clearly show that large-scale GREW training data is helpful for future gait recognition research. For the GEINet baseline, the scale of training data does not obviously influence the performance. The reason may be that the network architecture in GEINet has limited capability to learn from large data. TS-CNN uses a two-stream metric learning network structure and takes pairs of GEIs as inputs, which may be less suffered from over-fitting. Therefore, its Rank-1 accuracy slightly increases from 9.50\% to 13.55\%.
Model-based baselines are not sensitive to training data scales due to inferior accuracy.

\begin{figure}[htbp]
\centering
\includegraphics[width=0.8\linewidth]{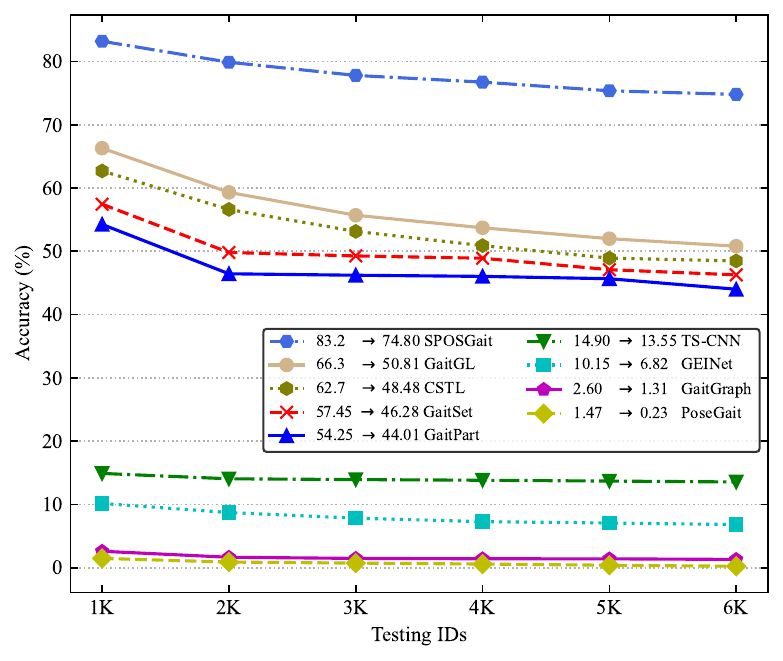}
\vspace{-5mm}
\caption{Rank-1 accuracy (\%) with different identities in the test set.
The legends show performance changes from 1K to 6K test data.
}
\label{fig:different_identities_in_test_set}
\end{figure}

\begin{figure*}[htbp]
\centering
\includegraphics[width=0.8\linewidth]{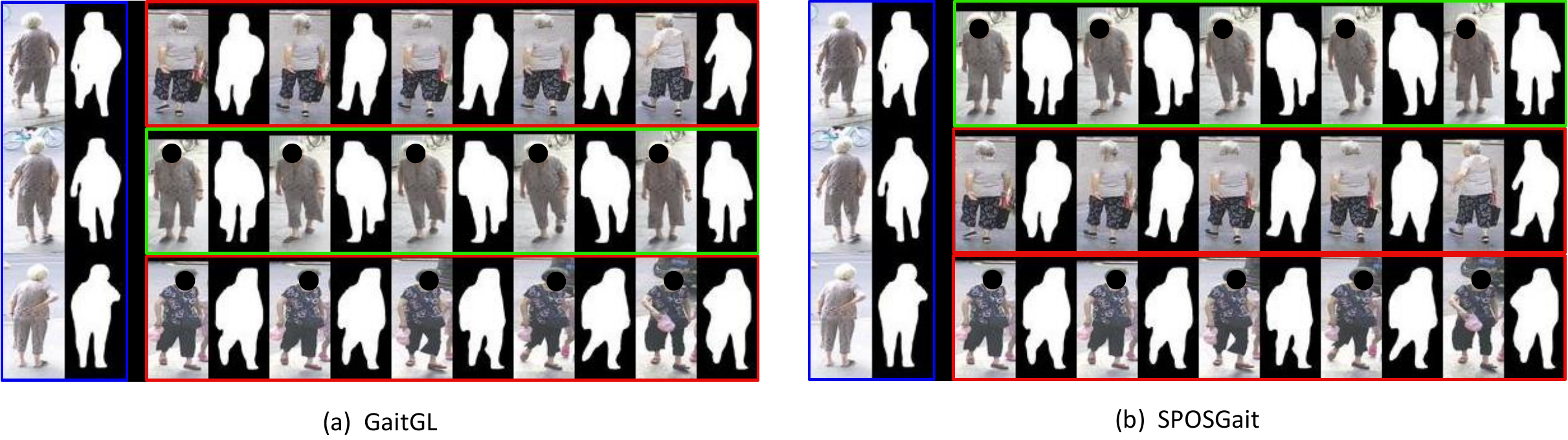}
\vspace{-4mm}
\caption{Sample results on the GREW with GaitGL and SPOSGait. The left part with \textcolor{blue} {blue} boxes shows probes (3 frames belong to the same sequence), while results with \textcolor{green} {green} and \textcolor{red}{red} boxes are true positives and false positives, respectively. Note that only silhouettes are used for gait recognition, and RGB images are just used for visualization.
}
\label{fig:Sample_results}
\vspace{-4mm}
\end{figure*}

\noindent{\bf Accuracy with Increasing Test Identities.}
A sufficient test set is essential for evaluating the performance of the gait recognizer. In this experiment, we study the relationship between the search space scale and the Rank-1 accuracy as shown in Figure \ref{fig:different_identities_in_test_set}. When the test identities increase from 1K to 6K, almost all approaches suffer from accuracy degradation. More specifically, SPOSGait scores 83.20\% Rank-1 on 1K test identities but decreases to 79.88\% when the test size is doubled. When the subjects increase to 6K, the precision degradation of SPOSGait is near 10\%. With increasing identities in the gallery, the possibility of inter-subject appearance similarity becomes higher, so recognizing certain identities by top retrieval is more challenging.
Evaluation results on other baselines come to the same conclusion.

\begin{figure}[htbp]
\centering
\includegraphics[width=0.8\linewidth]{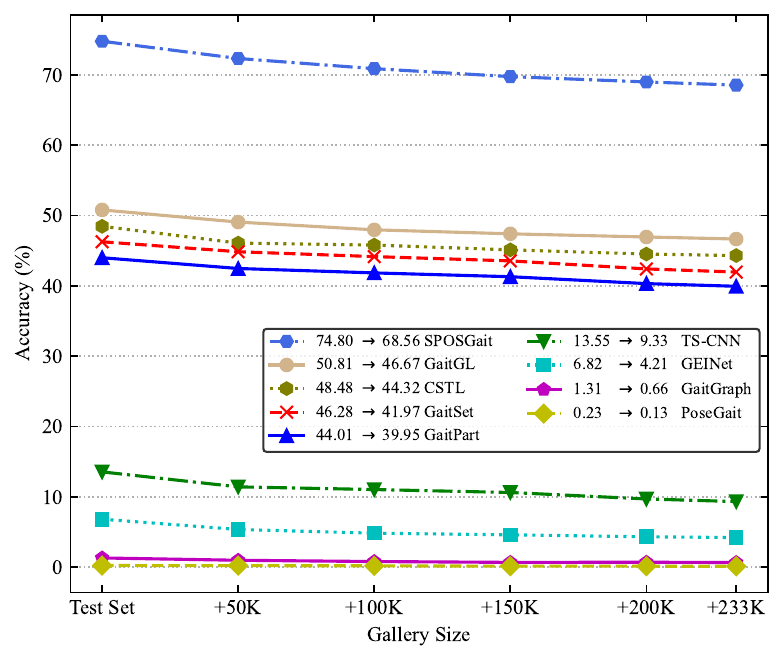}
\vspace{-4mm}
\caption{Rank-1 accuracy (\%) with increasing gallery size. Different distractor scales are added. Legends show performance changes from \texttt{test} to \texttt{test} + \texttt{distractor}.}
\label{fig:different_gallery_size}
\vspace{-4mm}
\end{figure}

\noindent{\bf Accuracy with Distractor Set.}
In gait applications, the gallery size may be very large considering numerous unrelated identities.
We add the constructed distractor set into the gallery to investigate this practice setting.
As shown in Figure \ref{fig:different_gallery_size}, by enlarging the gallery with a distractor set, most approaches obtain lower recognition scores. When all 233K distractor sequences are involved, Rank-1 of SPOSGait decreases to 68.56\%. Accuracy with the distractor set shows the necessity of the GREW benchmark again.


\begin{table}[htbp]
\vspace{-4mm}
\caption{Rank-1 accuracy (\%) on carrying and dressing attributes.
Subsets of probe (sequences with the corresponding attribute) are chosen to perform gait recognition.
For evaluation with dressing, \emph{All} means gait probe and gallery are paired without attention to any clothing style. \emph{Short/Long} refers to short/long-wearing in both upper and lower body.
}
\centering
\begin{center}{
\smallskip
\begin{tabular}{c|c|c|c}
\hline

Carrying & Rank-1 &  Dressing  & Rank-1 \\

\hline\hline
None & 62.26 & All & 74.80  \\
Backpack & 74.46 & Short & 76.92  \\
Shoulder bag & 73.35 & Long & 73.79  \\
Handbag & 79.45 & Skirt & 77.48  \\
Lift-stuff & 75.08 & - & -  \\

\hline
\end{tabular}}
\end{center}

\label{table:performance_on_carrying_and_dressing_attributes}
\vspace{-4mm}
\end{table}
\begin{figure}[htbp]
\centering
\includegraphics[width=0.8\linewidth]{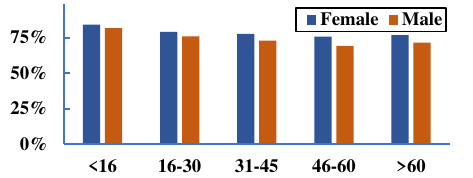}
\vspace{-4mm}
\caption{Rank-1 accuracy (\%) on gender and age group attributes.}
\label{fig:performance_on_age_and_gender}
\vspace{-6mm}
\end{figure}

\subsubsection{Performance on Different Attributes}

This section investigates the performance variations of gait recognition between different attributes, including gender, age group, carrying, and dressing. 
We adopt SPOSGait as the recognition approach.

The Rank-1 accuracy for gender and age group is illustrated in Figure \ref{fig:performance_on_age_and_gender}. According to the results, for most age groups, gait recognition performance in females is always better than that of males. We argue that females contain more different variations such as wearing and hairstyle, which may be helpful for individual recognition by gait silhouettes. For results on different age groups, one can find that the performance remains consistent, which highlights the exceptional efficacy and robustness of our approach. The attribute results on carrying and dressing are shown in Table \ref{table:performance_on_carrying_and_dressing_attributes}. Compared with normal walking (\ie \emph{None}), the Rank-1 accuracy significantly improves for other carrying scenarios, with the highest accuracy of 79.45\% when the subject is carrying a handbag. Similarly, the performance of SPOSGait varies with different dressing styles, with the highest accuracy observed for subjects wearing skirts.

\subsubsection{Sample results}
Figure \ref{fig:Sample_results} provides several sample results on the GREW \texttt{test} set, which are performed by GaitGL and SPOSGait. The GaitGL method encounters difficulties in generating accurate predictions, leading to erroneous results. Conversely, the SPOSGait demonstrates a higher degree of precision and effectiveness, consistently yielding correct predictions.

\section{Discussion and Conclusion}
\label{Discussion}
\subsection{Discussion}
During construction of the GREW benchmark, \emph{privacy and bias problems are our first concern}. To protect privacy, \emph{only silhouettes, flow, and human poses would be utilized and released}. We \emph{provide strict access for applicants who sign the license, and try our best to guarantee it for research purposes only}. For dataset bias, the GREW has a balanced gender distribution, while some attributes (\eg race, age group, dressing) are inevitably biased due to capture location and time. Since our dataset is large-scale and diverse, one can sample balanced data to train models with less bias. Besides, recent de-bias research in the biometrics community \cite{RFW,gong2020jointly,wang2020mitigating} may also alleviate this problem.


\subsection{Conclusion}
This paper makes the first step to large-scale gait recognition in the wild, to the best of our knowledge.
Firstly, the GREW dataset contains 128K sequences of 26K subjects with rich attribute variations from flexible data.
Secondly, we manually annotate thousands of hours of streams from hundreds of cameras, resulting in 14M boxes with automatic silhouettes and human poses. Moreover, 233K distractor set sequences are collected for practical evaluation. 
Lastly, we introduce a strong and highly robust method based on NAS for gait recognition and conduct comprehensive baselines to quantitatively analyze the challenges in unconstrained gait recognition, deriving in-depth and constructive insights. 


\small
\bibliographystyle{ieee}
\bibliography{egbib}

\end{document}